%% file: egpaper_final.tex
\documentclass[10pt,twocolumn,letterpaper]{article}

\usepackage{iccv}
\iccvfinalcopy
\usepackage{times}
\usepackage{environ} 
\usepackage{graphicx}
\usepackage{amsmath}
\usepackage{amssymb}
\usepackage{url} 
\usepackage{booktabs}
\usepackage{amsfonts}       
\usepackage{nicefrac}       
\usepackage{microtype}      
\usepackage{cite}
\usepackage{array}
\usepackage{pifont}
\usepackage{capt-of,etoolbox}
\usepackage{multirow}
\usepackage{hhline}
\usepackage{bbding}
\usepackage{algorithm}
\usepackage{algorithmic}
\usepackage{boldline}
\usepackage{braket}
\usepackage{enumitem}
\usepackage{wrapfig,lipsum}
\usepackage[12pt]{moresize}
\usepackage[table,xcdraw]{xcolor}
\NewEnviron{NORMAL}{%
    \scalebox{0.88}{$\BODY$} 
} 
\usepackage[T1]{fontenc}

\usepackage[pagebackref=true,breaklinks=true,letterpaper=true,colorlinks,bookmarks=false]{hyperref}

\usepackage[capitalize]{cleveref}
\crefname{section}{Sec.}{Secs.}
\Crefname{section}{Section}{Sections}
\Crefname{table}{Table}{Tables}
\crefname{table}{Tab.}{Tabs.}

\def\iccvPaperID{5766} 

\ificcvfinal\fi

\begin{document}

\title{Meta-Optimization for Higher Model Generalizability in Single-Image Depth Prediction}

\author{Cho-Ying Wu\quad Yiqi Zhong\quad Junying Wang\quad Ulrich Neumann \\ \\
University of Southern California\\
}

\makeatletter
\let\@oldmaketitle\@maketitle
\renewcommand{\@maketitle}{\@oldmaketitle
\vspace{-18pt}
\centering\includegraphics[width=0.83\linewidth]{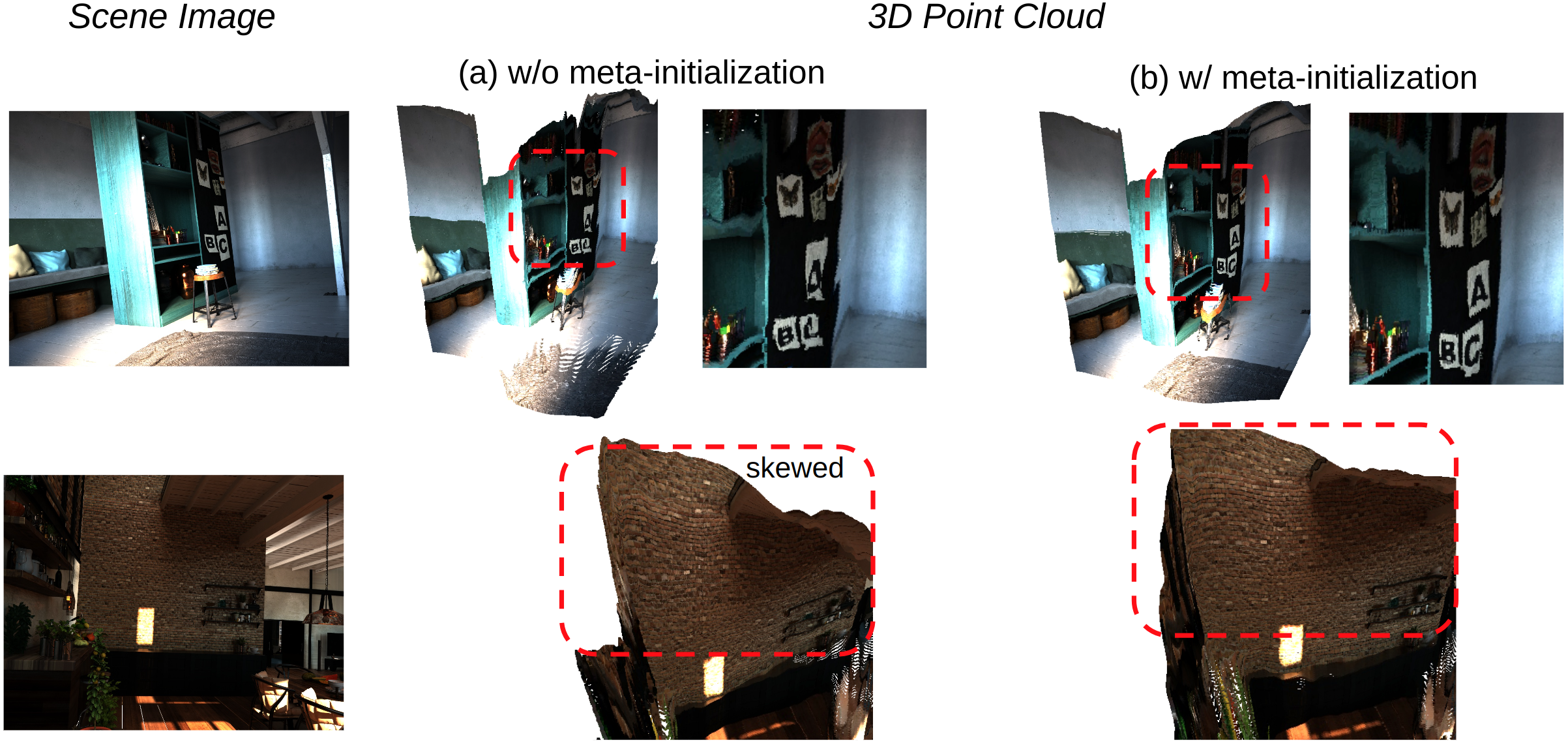}
\vspace{-4pt}
\captionof{figure}{\textbf{Geometry structure comparison in 3D point cloud view.} We back-project the predicted depth maps from images into textured 3D point cloud to show the geometry. The proposed Meta-Initialization has better domain generalizability that leads to more accurate depth prediction hence better 3D structures. (zoom in for the best view).  \\}
\label{pointcloud}}
  
\maketitle

\input{00_abstract}
\input{01_intro}
\input{02_related}
\input{03_methods}

\input{04_exp}

\input{05_conclusion}

{\small
\bibliographystyle{ieee_fullname}
\bibliography{egbib}
}

\end{document}


\renewcommand{\theequation}{S\arabic{equation}}
\renewcommand{\thefigure}{S\arabic{figure}}
\renewcommand{\thetable}{S\arabic{table}}
\renewcommand\thesection{\Alph{section}}
\renewcommand\thesubsection{\thesection.\Alph{subsection}}

\begin{figure*}[bt!]
    \centering
    \includegraphics[width=0.98\linewidth]{figures/samples_replica.png}
    \includegraphics[width=0.98\linewidth]{figures/samples_hm3d.png}
    \includegraphics[width=0.98\linewidth]{figures/samples_hypersim.png}
    \includegraphics[width=0.98\linewidth]{figures/samples_VA.png}
    \includegraphics[width=0.98\linewidth]{figures/samples_NYUv2.png}
    \vspace{-10pt}
    \caption{\textbf{Data samples for all adopted datasets.} Replica \cite{straub2019replica} has lower scene variety. The two shown sequences use the same structure with only some different object arrangements. HM3D \cite{ramakrishnan2021habitat} has higher scene variety for indoor spaces with different usages, but textures in their meshes are coarse. Hypersim \cite{roberts2020hypersim} contains much higher scene variety with many designed spaces, from small-size to large-size rooms, for different purposes. VA \cite{wu2022toward} comprises more variations in camera viewing directions for studying challenging scene structures. NYUv2 \cite{silberman2012indoor} contains real captures but mainly for smaller room spaces, and camera viewing directions mostly parallel to the grounds.}
    \label{data_samples}
     \vspace{-10pt}
\end{figure*}
\clearpage

\begin{Center}
{\LARGE \textbf{Supplementary Materials}}
\end{Center}

\section{Overview}
\label{overview}

\begin{itemize}[leftmargin=*,topsep=-0pt,itemsep=-0.0ex]
  \item In Sec.~\ref{sec:samples} we provide examples of each dataset we adopt.
  \item In Sec.~\ref{sec:relevant} we provide more explanations and analysis to depth-relevant and depth-irrelevant features.
  \item In Sec.~\ref{sec:relation} we extend the discussion in main paper Sec. 4.
  \item In Sec.~\ref{sec:metrics_formulation} we show the formula of depth evaluation metrics and organize TL;DR for terms used in the work.
  \item In Sec.~\ref{sec:pseudo} we display the pseudo-code for direct supervised learning and meta-learning under our fine-grained task setting.
  \item In Sec.~\ref{sec:more_results} we provide more studies on different learning strategies, compare with other cross-dataset evaluation works, plug meta-initialization into existing frameworks to validate meta-learning, show extensive qualitative depth map results, and show more quantitative and qualitative results for depth-supervised NeRF.
  \item In Sec.~\ref{sec:broader} we illustrate the broader impact related to this work.
  
\end{itemize}

\section{Data Samples}
Examples of all the adopted datasets and their features are shown in Fig. \ref{data_samples}.
\label{sec:samples}

\section{Depth-Relevant and Depth-Irrelevant Features}
\label{sec:relevant}
In Introduction and Section 3.1 of the main paper, we explain the division between depth-relevant and depth-irrelevant features: whether pixel color or appearance changes indicate depth changes. An example of the former is foreground object boundaries, where the color changes imply depth changes. By contrast, simple material textures or paintings are depth-irrelevant. We show an illustration in Fig. \ref{depth-relevant}.
\begin{figure}[h]
    \centering
    \includegraphics[width=0.35\linewidth]{figures/DepthRelevant.png}
    \vspace{-12pt}
    \caption{\textbf{Illustration of the division between depth-relevant and depth-irrelevant cues.}}
    \vspace{-4pt}
    \label{depth-relevant}
\end{figure}
In Fig. 3 of the paper, we show that meta-learning can induce better image-to-depth understanding and suppress depth-irrelevant features, such as flat areas on the textured carpet and clearer object boundaries in depth maps. For deeper insight, the experiment in paper Fig. 3 is trained on Replica, which contains only 18 environments, and some possess the same structures with minor arrangement changes (See Fig.~\ref{data_samples}). Limited scene variety makes it difficult for direct supervised learning to attain good or valid image-to-depth understanding, and thus it reaches inferior performance. Therefore, as shown in paper Fig. 3, it cannot suppress depth-irrelevant cues and reflect texture patterns in the depth maps. 

In contrast, meta-learning is good at few-shot or low-source learning in literature because of its learning nature. It can learn generalizable good or valid image-to-depth mappings from scenes with limited scene variety. Its dual-step optimization does not directly fit each training pair but uses a step size $\beta$ in Algorithm 1 (meta-initialization) to control how much the explored gradient updating direction is trusted. Thus, it avoids directly fitting each seen example and allows more exploration in the neighborhood of each solution point, achieving better image-to-depth understanding for higher depth accuracy. We show a comparison in Fig. \ref{explanation} for the effects of the $\beta$ parameter.

\begin{figure}[h]
    \centering
    \includegraphics[width=0.78\linewidth]{figures/explanation2.png}
    \vspace{-7pt}
    \caption{\textbf{Effects of different step size parameter $\beta$.} ConvNeXt-base architecture is used. We show that using a larger step size parameter $\beta=0.9$, the training becomes more similar to direct supervised learning that tries to fit each seen training sample, but may not fully explore the neighborhood for each intermediate solution and attain better image-to-depth mappings as meta-learning performs. See the main text.}
    \vspace{-8pt}
    \label{explanation}
\end{figure}

\section{How is fine-grained task related to other meta-learning studies?}
\label{sec:relation}

There are several previous findings on learning techniques or issues related to meta-learning. Here we discuss how those findings apply to fine-grained tasks.

\textbf{Relation to domain-agnostic task augmentation}. Domain-agnostic task augmentation is to densify sampled data points in each task to add robustness, such as label noise \cite{rajendran2020meta}, image transformation \cite{ni2021close}, and MetaMix \cite{yao2021improving}. Since our fine-grained task contains only one sample for every task, the domain-agnostic augmentation is reduced to data augmentation, where we can simply inject mild depth label noise and image transformation, such as left-right flip and color jittering, to create derivative samples associated with each fine-grained task.  

\textbf{Relation to task interpolation}. Unlike domain-agnostic augmentation, task interpolation (MLTI \cite{yao2021meta}) is investigated to remedy the requirements of large numbers of meta-training tasks. MLTI adopts the following interpolation method to augment tasks.
\begin{equation}
\vspace{-0pt}
     H_{cross} = \lambda H_i + (1-\lambda) H_j, \qquad Y_{cross} = \lambda Y_i + (1-\lambda) Y_j,
\label{mlti}
\vspace{-0pt}
\end{equation}
where subscript $i,j$ denote different tasks. $H$ represents intermediate features, $Y$ represents labels, and $\lambda$ is a weight to control the interpolation between the two tasks. Eq. \ref{mlti} may not be valid for our fine-grained tasks. For example, mixing features from different scenes does not lead to the same interpolation for depth ranges, i.e., overlaying near and far range scenes together can break local region dependency in each scene for depth prediction, and thus does not naturally indicate depth values in between. A feasible approach can be fusing geometry from one scene with texture from another, or performing 3D-aware scene interpolation, which not only interpolates intermediate depth ranges but also attends to geometry and appearance coherence.   

\textbf{Relation to meta-memorization and meta-overfitting}. Prior meta-learning studies \cite{ yin2020meta,rajendran2020meta,yao2021improving} point out that memorization and overfitting may also occur at the task level. However, they validate the claim on rather simple problems and settings such as toy sinusoidal regression and pose prediction using 15-class Pascal3D objects \cite{xiang2014beyond} for chair or sofa CAD models rendered on flat grounds. Their task complexity is relatively simple, and a network may easily memorize all tasks without generalizability, compared to our study on per-pixel real-valued depth prediction. Our fine-grained task setting shows that studying more practical and complex problems may be free from task memorization and overfitting. As shown in paper experiments, without using techniques such as meta-regularization and meta-augmentation to relieve the meta-memorization and meta-overfitting, our network can still serve as a good initialization in either intra-dataset or cross-dataset evaluation.

\section{Error Metrics Formula for Depth Evaluation and Term Dictionary}
\label{sec:metrics_formulation}
We provide formula for adopted depth evaluation metrics between prediction ($x_s$) and groundtruth ($y_s$), $\forall s \in S$, as follows.

(1) MAE: $\frac{1}{|S|} \sum_{s \in S} |x_s-y_s| $.

(2) AbsRel: $\frac{1}{|S|} \sum_{s \in S} \frac{|x_s-y_s|}{y_s} $.

(3) RMSE: $\sqrt{ \frac{1}{|S|} \sum_{s \in S} (x_i-y_i)^2}$.

(4) RMSE$_{log}$: $\sqrt{ \frac{1}{|S|} \sum_{s \in S} (log(x_i)-log(y_i))^2}$.

The above four are error metrics. The lower the better.

(5) Depth accuracy $\delta_i$:

\begin{equation}
\delta_i = \frac{\text{card}(\Big\{x_s: \max \{ \frac{x_s}{y_s},\frac{y_s}{x_s}\} < 1.25^i \Big\} )}{\text{card}(\{y_s)\}},
\label{ls_reform}
\end{equation}
where card(.) is the cardinality of a set. This is an accuracy metric. The higher the better.

\textbf{Variance}. We also observe that meta-initialization induces smaller variances on error  $|x_s-y_s|$. Error variance for Table 4 Hypersim$\to$Replica are \textit{DPT-hybrid}: 0.236, \textit{DPT-hybrid + Meta}: 0.201, \textit{DPT-large}: 0.218, \textit{DPT-large + Meta}: 0.199. This shows that meta-initialization tends to predict more structured depth in reasonable ranges, preventing jumpy depth that causes large errors.  

\textbf{Dictionary}. We provide quick explanations as TL;DR for terms used in the paper.

\begin{itemize}[leftmargin=*,topsep=-5pt,itemsep=-0.8ex]
\item \textbf{generalizability} refers to whether a pretrained model can generalize to unseen data and make reasonably good inferences. This work especially stresses generalizability to unseen data from different datasets.

\item \textbf{zero-shot cross-dataset inference} refers to training on $A$-dataset without any knowledge on $B$-dataset and making inference on $B-$dataset.

\item\textbf{scene variety} refers to variety of scene appearance and geometry (RGBD) pairs in a dataset.

\item\textbf{task} in meta-learning context contains a distribution to sample data from. Those data share similarities or affinity so that they can be grouped together.

\item\textbf{depth-relevant/ depth-irrelevant low-level cues} refer to whether pixel color or appearance changes as low-level cues indicate depth changes. An example of the former is foreground object boundaries. Simple material textures or paintings are depth-irrelevant.
\end{itemize}

\section{Pseudo-code}
\label{sec:pseudo}
We display pseudo-code for direct supervised learning and our fine-grained task meta-learning as follows. Our fine-grained task meta-learning only needs to adapt a few lines of codes in a conventional supervised learning framework to build bi-level optimization. With this simple plug-in, our fine-grained task meta-learning effectively learns better domain generalizability and higher geometry resolvability, as shown in the qualitative and quantitative evaluation.

\begin{lstlisting}[language=Python, caption=\textit{PyTorch-like} pseudo-code for direct supervised learning and our fine-grained task meta-learning]
# I: image as a minibatch
# D: depth groundtruth of I
def direct_supervised_learning(I,D):
    optimizer.zero_grad() # flush out gradient
    D_pred = model(I) # predict depth
    loss = criterion(D_pred,D) # calculate loss
    loss.backward() # back-prop
    optimizer.step() # update network
    
def meta_learning(I,D):
    meta_optimizer.zero_grad() # flush out gradient
    for step in range(L): # L-step
        inner_optimizer.zero_grad() # flush out gradient
        D_pred = inner_model(I) # predict depth
        loss = criterion(D_pred,D) # calculate loss
        loss.backward() # back-prop
        optimizer.step() # SGD-update inner-network
    for meta_param, inner_param = zip(meta_model.parameters(), inner_model.parameters()):
        # assign gradient as parameter difference
        meta_param.grad = meta_param - inner_param 
    meta_optimizer.step() # SGD-update meta-network

\end{lstlisting}

\section{More Results}
\label{sec:more_results}
\textbf{Comparison with simple pretraining on the same dataset}.
We first compare with \textit{simple pretraining} on the same dataset but with different learning schedules. 
The networks are pretrained by 5 epochs using larger and feasible learning rates of 0.001 and 0.0003 and a strong weight decay of 0.1. Then, the learned weights serve as initialization for the following supervised learning, whose setting is the same as in the main paper. The purpose is trying to examine whether a higher-level and smooth prior can be learned without using meta-learning. Then the same as meta-initialization, we use the learned weights as initialization for the second-stage supervised learning. We use ResNet50 and ConvNext-base and train/test on NYUv2. 
Results of different learning rates and weight decay are compared and shown in Table~\ref{table:add_pre}.
We find that larger learning rates and weight decay cannot learn a good prior but damage the performance. Besides, higher weight decay did not result in apparent positive effects. 
Note that "w/o Pretraining" simply uses the second-stage supervised learning. The entry "w/ Pretraining (lr=3x10$^{-4}$, wd=10$^{-2}$)" is equivalent with longer training for "w/o Pretraining" since its learning rate and weight decay match those used in the "w/o Pretraining." 
The results show that simple pretraining cannot learn a better prior. 
Thus, we resort to meta-learning with its advantages of higher model generalizability in literature.
\begin{table}[h]
\begin{center}
  \caption{\textbf{Comparison with simple pretraining strategy.} Adopted architecture, learning rate (lr), and weight decay (wd) are shown. The pretraining first uses a higher lr and wd of 0.001 and 0.1 to learn a smooth prior. We also experiment with different lr and wd for comparison. The learned weights are then used as initialization for the second-stage supervised learning. See text for the pretraining setting. The pretraining does not improve over baseline without this trick, and larger weight decay slightly degrades the performance.}
  \vspace{-8pt}
  \footnotesize
  \label{table:add_pre}
  \begin{tabular}[c]
  {
  p{5.5cm}<{\arraybackslash}
  p{0.6cm}<{\centering\arraybackslash}
  p{0.7cm}<{\centering\arraybackslash}
  p{0.7cm}<{\centering\arraybackslash}
  p{0.5cm}<{\centering\arraybackslash}
  p{0.5cm}<{\centering\arraybackslash}
  p{0.5cm}<{\centering\arraybackslash}}
  \hlineB{2}
  
      \multicolumn{1}{|c|}{\cellcolor[HTML]{99CCFF} NYUv2}  & \cellcolor[HTML]{FAE5D3} MAE & \cellcolor[HTML]{FAE5D3} AbsRel & \cellcolor[HTML]{FAE5D3} RMSE &  \multicolumn{1}{|c|}{\cellcolor[HTML]{D5F5E3} $\delta_1$} &  \multicolumn{1}{|c|}{\cellcolor[HTML]{D5F5E3} $\delta_2$} &  \multicolumn{1}{|c|}{\cellcolor[HTML]{D5F5E3} $\delta_3$} \\
    \hline
     \multicolumn{7}{c}{\cellcolor[HTML]{FFFE65}ResNet50} \\
      w/o Pretraining  & 0.345 & 0.131 & 0.480 & 83.6 & 96.4 & 99.0 \\
      w/ Pretraining (lr=10$^{-3}$, wd=10$^{-1}$)  & 0.362 & 0.138 & 0.500 & 82.9 & 95.6 & 97.9 \\
      w/ Pretraining (lr=3x10$^{-4}$, wd=10$^{-1}$) & 0.347 & 0.132 & 0.481 & 83.5 & 96.4 & 99.0 \\
      w/ Pretraining (lr=3x10$^{-4}$, wd=10$^{-2}$) & 0.345 & 0.133 & 0.480 & 83.6 & 96.4 & 99.0 \\
      w/ Meta-Initialization  & \textbf{0.325} & \textbf{0.122} & \textbf{0.454} & \textbf{85.4} & \textbf{96.8} & \textbf{99.3} \\
      \hline
      \multicolumn{7}{c}{\cellcolor[HTML]{FFFE65}ConvNeXt-base} \\
      w/o Pretraining & 0.273 & 0.101 & 0.394 & 89.4 & 97.9 & \textbf{99.5} \\
      w/ Pretraining (lr=10$^{-3}$, wd=10$^{-1}$) & 0.288 & 0.109 & 0.414 & 87.5 & 97.5 & 99.4 \\
      w/ Pretraining (lr=3x10$^{-4}$, wd=10$^{-1}$) & 0.276 & 0.103 & 0.397 & 89.2 & 97.9 & \textbf{99.5} \\
      w/ Pretraining (lr=3x10$^{-4}$, wd=10$^{-2}$) & 0.274 & 0.101 & 0.395 & 89.3 & 97.9 & \textbf{99.5} \\
      w/ Meta-Initialization & \textbf{0.266} & \textbf{0.099} & \textbf{0.387} & \textbf{89.8} & \textbf{98.1} & \textbf{99.5} \\
    \hlineB{2}
    \hline
  \end{tabular}
  \vspace{-8pt}
\end{center}
\end{table}

\textbf{Comparison with gradient accumulation}.
We next compare with gradient accumulation. Setting-1:  Similar to the prior-learning stage, gradients are accumulated for 4 iterations and then used to update network parameters once. This resembles taking off the inner exploration and inner optimizer in meta-learning. We train this strategy for 5 epochs with a learning rate of 0.0012, 4x by its base learning rate since we accumulate gradients for 4 iterations. Then we used the learned weights as initialization for the following standard supervised learning whose hyperparameters are the same as in the main paper. Setting-2: We adopt a single-stage approach, which does not require the prior-learning stage, and simply use the gradient accumulation trick in the standard supervised learning. We also accumulate gradients for 4 iterations and use a learning rate of 0.0012. The rest hyperparameters are intact.
We again use ResNet50 and ConvNext-base and train/test on NYUv2. Results are shown in Table~\ref{table:add_gd}.
From the table, we empirically find both settings do not improve the results of "Base", which is standard supervised learning without any add-on methods. We think this is because gradient accumulation has effects of using large batch size, which has a higher risk to overfit training data by converging to poor local optima, due to the reduction of stochasticity in the gradient updates ~\cite{keskarlarge}.
We also try accumulation for 8 iterations and scale learning rates accordingly. On NYUv2 using ResNet50, RMSE for Setting-1 and Setting-2 are 0.493 and 0.498, which are on par and slightly worse than accumulation steps of 4.
In summary, this experiment justifies that the improvements come from the bilevel-fashion of meta-learning that helps learn a good prior.

\begin{table}[h]
\begin{center}
  \caption{\textbf{Comparison with gradient accumulation.} Two settings in comparison are described in Sec.~\ref{sec:more_results}. "Base" refers to using standard supervised learning without any add-on methods. Empirically we find gradient accumulation does not improve results but degrades performance a little.}
  \vspace{-8pt}
  \footnotesize
  \label{table:add_gd}
  \begin{tabular}[c]
  {
  p{2.5cm}<{\arraybackslash}
  p{0.6cm}<{\centering\arraybackslash}
  p{0.7cm}<{\centering\arraybackslash}
  p{0.7cm}<{\centering\arraybackslash}
  p{0.5cm}<{\centering\arraybackslash}
  p{0.5cm}<{\centering\arraybackslash}
  p{0.5cm}<{\centering\arraybackslash}}
  \hlineB{2}
  
      \multicolumn{1}{|c|}{\cellcolor[HTML]{99CCFF} NYUv2}  & \cellcolor[HTML]{FAE5D3} MAE & \cellcolor[HTML]{FAE5D3} AbsRel & \cellcolor[HTML]{FAE5D3} RMSE &  \multicolumn{1}{|c|}{\cellcolor[HTML]{D5F5E3} $\delta_1$} &  \multicolumn{1}{|c|}{\cellcolor[HTML]{D5F5E3} $\delta_2$} &  \multicolumn{1}{|c|}{\cellcolor[HTML]{D5F5E3} $\delta_3$} \\
    \hline
     \multicolumn{7}{c}{\cellcolor[HTML]{FFFE65}ResNet50} \\
      Base & 0.345 & 0.131 & 0.480& 83.6& 96.4& 99.0\\
      Setting-1  & 0.349 & 0.133 & 0.487 & 83.4 & 96.3 & 99.0 \\
      Setting-2  & 0.353 & 0.134 & 0.493 & 83.1 & 96.1 & 98.8 \\
      Meta-Initialization  & \textbf{0.325} & \textbf{0.122} & \textbf{0.454} & \textbf{85.4} & \textbf{96.8} & \textbf{99.3} \\
      \hline
      \multicolumn{7}{c}{\cellcolor[HTML]{FFFE65}ConvNeXt-base} \\
      Base & 0.273 & 0.101 & 0.394& 89.4& 97.9& \textbf{99.5}\\
      Setting-1 & 0.277 & 0.103 & 0.399 & 89.1 & 97.8 & 99.4 \\
      Setting-2 & 0.279 & 0.105 & 0.406 & 88.9 & 97.7 & 99.4 \\
      Meta-Initialization & \textbf{0.266} & \textbf{0.099} & \textbf{0.387} & \textbf{89.8} & \textbf{98.1} & \textbf{99.5} \\
    \hlineB{2}
    \hline
  \end{tabular}
  \vspace{-8pt}
\end{center}
\end{table}

\textbf{Additional comparison to other works}.
Few recent works are related to cross-dataset evaluation for indoor depth, including unsupervised domain adaptation \cite{zheng2018t2net,chen2019crdoco,zhao2019geometry} and semi-supervised \cite{zhao2020domain}. However, they require access to target domain for adaptation.
Another work S2R-DepthNet also learns generalizable depth estimation \cite{Chen2021S2RDepthNet} and reports \textit{zero-shot} cross dataset performance.

We further compare to an unsupervised domain adaptation method T$^2$Net~\cite{zheng2018t2net}, semi-supervised domain adaptation method ARC~\cite{zhao2020domain} and S2R-DepthNet~\cite{Chen2021S2RDepthNet}. These works adopt the setting SUNCG$\to$NYUv2.
We follow training scripts by S2R-DepthNet and show results in Table ~\ref{table:add1}.


\begin{table}[h]
\begin{center}
  \caption{\textbf{Cross-Dataset comparison to other works.} Numbers are taken from \cite{Chen2021S2RDepthNet}.}
  \vspace{-5pt}
  \small
  \label{table:add1}
  \begin{tabular}[c]
  {
  p{2.9cm}<{\arraybackslash}
  p{0.8cm}<{\centering\arraybackslash}
  p{0.8cm}<{\centering\arraybackslash}
  p{0.6cm}<{\centering\arraybackslash}
  p{0.6cm}<{\centering\arraybackslash}
  p{0.6cm}<{\centering\arraybackslash}}
  \hlineB{2}
  
      \multicolumn{1}{|c|}{\cellcolor[HTML]{99CCFF} SUNCG$\to$NYUv2}  &  \cellcolor[HTML]{FAE5D3} AbsRel & \cellcolor[HTML]{FAE5D3} RMSE &  \multicolumn{1}{|c|}{\cellcolor[HTML]{D5F5E3} $\delta_1$} &  \multicolumn{1}{|c|}{\cellcolor[HTML]{D5F5E3} $\delta_2$} &  \multicolumn{1}{|c|}{\cellcolor[HTML]{D5F5E3} $\delta_3$} \\
    \hline
    T$^2$Net ~\cite{zheng2018t2net} & 0.203 & 0.738 & 67.0 & 89.1 & 96.6 \\
    ARC~\cite{zhao2020domain} & 0.186 & 0.710 & 71.2 & 91.7 & 97.7\\
      S2R-DepthNet  ~\cite{Chen2021S2RDepthNet} & 0.196 & 0.662 & 69.5 & 91.0 & 97.2 \\
      Our Meta-Initialization & \textbf{0.177} & \textbf{0.635} & \textbf{72.8} & \textbf{92.8} & \textbf{97.8} \\
      \hline
    \hlineB{2}
  \end{tabular}

  \vspace{-5pt}
\end{center}
\end{table}

\textbf{Additional results for adding meta-initialization to dedicated depth estimation architecture}.
In addition to Table 4 in the paper, we provide more experiments using other existing dedicated architecture for monocular depth estimation, including AdaBins~\cite{bhat2021adabins} and GLPDepth~\cite{kim2022global}. Following paper Table 4\footnote{In paper Table 4, redundant "our" are added. "+Meta" specifies plugging our proposed meta-initialization into those base training architecture.}, we train on Hypersim and test on Replica and NYUv2 and show results in Table ~\ref{table:cross-dataset-dedicated-add}. The tables in paper and here are run using official implementation. They both display that meta-initialization consistently improves performance using the dedicated architecture for single-image depth estimation.

\begin{table}[h]
\begin{center}
  \caption{\textbf{Extended comparison of zero-shot cross-dataset evaluation for dedicated depth estimation architecture.} }
  \vspace{-4pt}
  \footnotesize
  \label{table:cross-dataset-dedicated-add}
  
  \begin{tabular}[c]
  {
  p{2.9cm}<{\arraybackslash}|
  p{0.7cm}<{\centering\arraybackslash}|
  p{0.8cm}<{\centering\arraybackslash}|
  p{0.7cm}<{\centering\arraybackslash}|
  p{0.7cm}<{\centering\arraybackslash}|
  p{0.7cm}<{\centering\arraybackslash}|
  p{0.7cm}<{\centering\arraybackslash}}
  \hlineB{2}
  
       \multicolumn{1}{|c|}{\cellcolor[HTML]{99CCFF} Hypersim $\to$ Replica} & \cellcolor[HTML]{FAE5D3} MAE & \cellcolor[HTML]{FAE5D3} AbsRel & \cellcolor[HTML]{FAE5D3} RMSE & \cellcolor[HTML]{D5F5E3} $\delta_1$ & \cellcolor[HTML]{D5F5E3} $\delta_2$ & \cellcolor[HTML]{D5F5E3} $\delta_3$ \\
    \hline
    AdaBins \cite{bhat2021adabins} & 0.395 & 0.210 & 0.564 & 70.4 & 88.1 & 94.7 \\
    AdaBins+Meta  & \textbf{0.377} & \textbf{0.198} & \textbf{0.541} & \textbf{71.6} & \textbf{89.2} & \textbf{95.5}\\
    \hline
    GLPDepth \cite{kim2022global} & 0.352 & 0.191 & 0.522 & 73.0 & 90.9& 96.5\\
    GLPDepth+Meta & \textbf{0.337} & \textbf{0.180} & \textbf{0.498} & \textbf{74.4} & \textbf{92.2}& \textbf{96.8}\\
    \hlineB{2}
    \hline
  
   \multicolumn{1}{|c|}{\cellcolor[HTML]{99CCFF} Hypersim $\to$ NYUv2} & \cellcolor[HTML]{FAE5D3} MAE & \cellcolor[HTML]{FAE5D3} AbsRel & \cellcolor[HTML]{FAE5D3} RMSE & \cellcolor[HTML]{D5F5E3} $\delta_1$ & \cellcolor[HTML]{D5F5E3} $\delta_2$ & \cellcolor[HTML]{D5F5E3} $\delta_3$ \\
    \hline
    AdaBins \cite{bhat2021adabins} & 0.469 & 0.188&0.642 & 72.6 &91.2 &96.6 \\
    AdaBins+Meta  & \textbf{0.448}& \textbf{0.175} & \textbf{0.625} & \textbf{74.0} & \textbf{92.6} & \textbf{97.4}\\
    \hline
      GLPDepth \cite{kim2022global} & 0.438 & 0.169 & 0.604 & 75.3 & 93.9 & 98.2 \\
      GLPDepth+Meta  & \textbf{0.414} & \textbf{0.158} & \textbf{0.583} & \textbf{77.9} & \textbf{94.3} & \textbf{98.3} \\
      
    \hlineB{2}
    \hline
  \end{tabular}
  \vspace{-5pt}
\end{center}
\end{table}

\textbf{Qualitative results on zero-shot cross-dataset evaluation.} Following the quantitative comparison in Table 3 and Table 4 in the main paper, we show qualitative comparisons to examine zero-shot cross-dataset evaluation. In Fig. \ref{table:extra-qualitative-hm3d}, we use ConvNeXt-Base as the backbone network, train on HM3D and make inferences on Replica and VA, and compare between using meta-initialization and without meta-initialization. In Fig. \ref{table:extra-qualitative-hpsim}, we use the dedicated depth estimation architecture, DPT-large, train on Hypersim and make inferences on NYUv2 and Replica, and also compare between using meta-initialization and without meta-initialization.

In both Fig. \ref{table:extra-qualitative-hm3d} and \ref{table:extra-qualitative-hpsim}, meta-initialization induces clearer depth shapes and outlines with less irregularity. The results show that on the challenging zero-shot cross-dataset evaluation, meta-initialization can learn higher model generalizability that transfers knowledge from synthetic datasets (HM3D and Hypersim) to more challenging and higher quality synthetic (VA) or real data (NYUv2) and estimates accurate depth shapes for them.

\textbf{Depth-supervised NeRF}. In the main paper Sec. 4.4 we train NeRF \footnote{Specifically, we use high-performing instant-ngp \cite{muller2022instant} implementation (https://github.com/NVlabs/instant-ngp)} with supervision by depth predicted from our meta-initialization strategy using ConvNeXt-Base backbone. To convert from depth (z-value) to ray distance in a pinhole camera model, we do the following conversion.
\begin{equation}
\vspace{-1pt}
     distance = depth \times \sqrt{1+(\frac{x-c_x}{f_x})^2+(\frac{y-c_y}{f_y})^2},
\label{supervised}
\vspace{-2pt}
\end{equation}
where $c_x$ and $c_y$ are principal point coordinates, and $f_x$ and $f_y$ are focal lengths.

\begin{table}[h]
\begin{center}
 \setlength{\abovecaptionskip}{3pt} 
  \caption{\textbf{More results on depth-supervised NeRF.} We test on Replica 'room-0', 'room-1', room-2', 'office-0', 'office-1', and 'office-2' environments. We train a NeRF on each environment with 180 views. The comparison between using depth from meta-initialization and w/o meta-initialization for supervision is drawn. PSNR and SSIM are image quality metrics; the higher, the better.}
  \small
  \label{nerf_metrics}
  \begin{tabular}[c]
  {|
  p{1.5cm}<{\centering\arraybackslash}|
  p{1.5cm}<{\centering\arraybackslash}|
  p{1.5cm}<{\centering\arraybackslash}|
  p{1.5cm}<{\centering\arraybackslash}|
  p{1.5cm}<{\centering\arraybackslash}|}
  \hline
  & \multicolumn{2}{c}{\cellcolor[HTML]{99CCFF}w/o meta-initialization} & \multicolumn{2}{c}{\cellcolor[HTML]{F7BCDC}w/ meta-initialization} \\
       Environment  &  PSNR &  SSIM &  PSNR &  SSIM  \\
    \hline
       Room-0 & 29.988 & 0.8184 & \textbf{30.920} & \textbf{0.8373} \\
       Room-1 & 34.547 & 0.9279 & \textbf{34.871} & \textbf{0.9305}\\ 
       Room-2 & 36.680 & 0.9560 & \textbf{37.460} & \textbf{0.9609} \\ 
       Office-0 & 38.674 & 0.9629 & \textbf{39.290} & \textbf{0.9680} \\
       Office-1 & 36.196 & 0.9427 & \textbf{36.867} & \textbf{0.9460} \\
       Office-2 & 42.648 & 0.9638 & \textbf{42.665} & \textbf{0.9646}  \\
    \hline
  \end{tabular}
  \vspace{-11pt}
\end{center}
\end{table}

We show more quantitative comparisons for depth-supervised NeRF in Table~\ref{nerf_metrics} on Replica. Each is trained with 180 views along with losses for pixel color and distances, as described in the main paper Sec.4.4. 
The use of meta-initialization consistently outperforms the baselines, without meta-initialization, in terms of image quality metrics. More qualitative comparisons are displayed in Fig.~\ref{nerf_rendering}.

\section{Broader Impact} 
\label{sec:broader}
The research focuses on using gradient-based meta-learning to improve monocular depth estimation performance. As monocular depth can be applied in indoor AR/VR creation and interaction, robot navigation, and learning 3D representations for general purposes, the proposed method can be a part of a training convention that facilitates depth estimation to attain each goal for each application, especially fulfill the purpose of in-the-wild robustness. 

\textbf{Ethical considerations}: This work studies how to improve model generalizability by meta-learning. The advantage this work brings about is better indoor depth estimation technology for applications such as AR/VR, gaming systems, or real estate demonstrations. Depth or geometric data are less sensitive since it provides only shape outlines that are less identifiable and do not leak personal information seriously.








\clearpage
\begin{figure}[h]
    \centering
    \makebox[\textwidth][c]{\includegraphics[width=1.35\linewidth]{figures/qual_extra_re_1.png}}
    \makebox[\textwidth][c]{\includegraphics[width=1.35\linewidth]{figures/qual_extra_re_2.png}}
    \makebox[\textwidth][c]{\includegraphics[width=1.35\linewidth]{figures/qual_extra_re_3.png}}
    \vspace{-10pt}
    \caption{\textbf{Qualitative comparison on cross-dataset inference using ConvNeXt-Base.} Highlighted areas show the differences. Zoom in for the best view.}
    \vspace{-1pt}
    \label{table:extra-qualitative-hm3d}
\end{figure}

\begin{figure}[h]
    \centering
    \makebox[\textwidth][c]{\includegraphics[width=1.35\linewidth]{figures/qual_extra_re_4.png}}
    \makebox[\textwidth][c]{\includegraphics[width=1.35\linewidth]{figures/qual_extra_re_5.png}}
    \makebox[\textwidth][c]{\includegraphics[width=1.35\linewidth]{figures/qual_extra_re_6.png}}
    \vspace{-10pt}
    \caption{\textbf{Qualitative comparison on cross-dataset inference using DPT-large \cite{Ranftl2021}.} Highlighted areas show the differences. Zoom in for the best view.}
    \vspace{-1pt}
    \label{table:extra-qualitative-hpsim}
\end{figure}

\begin{figure}[h]
    \centering
    \includegraphics[width=1.0\linewidth]{figures/nerf-1.png}
    \includegraphics[width=1.0\linewidth]{figures/nerf-2.png}
    \vspace{-10pt}
    \caption{\textbf{Image quality comparison for NeRF rendering.} We show the quality metrics (the higher the better) under each image. Zoom in for the best view.}
    \label{nerf_rendering}
\end{figure} 

\clearpage
\newpage

\small
\bibliographystyle{ieee_fullname}
\bibliography{egbib}

%% file: 00_abstract.tex
\begin{abstract}
\vspace{-20pt}
Model generalizability to unseen datasets, concerned with in-the-wild robustness, is less studied for indoor single-image depth prediction. We leverage gradient-based meta-learning for higher generalizability on zero-shot cross-dataset inference. Unlike the most-studied image classification in meta-learning, depth is pixel-level continuous range values, and mappings from each image to depth vary widely across environments. Thus no explicit task boundaries exist. We instead propose fine-grained task that treats each RGB-D pair as a task in our meta-optimization. We first show meta-learning on limited data induces much better prior (max +29.4\%). Using meta-learned weights as initialization for following supervised learning, without involving extra data or information, it consistently outperforms baselines without the method. Compared to most indoor-depth methods that only train/ test on a single dataset, we propose zero-shot cross-dataset protocols, closely evaluate robustness, and show consistently higher generalizability and accuracy by our meta-initialization. The work at the intersection of depth and meta-learning potentially drives both research streams to step closer to practical use.
\end{abstract}

%% file: 01_intro.tex
\section{Introduction}
\label{sec:intro}

Much research attempts to learn geometry of scenes, representing in depth maps, from single images to fulfill physical indoor applications with depth such as collision detection \cite{flacco2012depth, nascimento2020collision,wu2020geometry}, robot navigation \cite{tai2018socially,tan2022depth,irshad2021hierarchical}, grasping \cite{irshad2021hierarchical,viereck2017learning,schmidt2018grasping}, human verification or interaction~\cite{wu2022cross,wu2021synergy,wu2018occluded,wu2016occlusion,wu2021scene} or it benefits learning good 3D representations for novel view synthesis using depth supervision \cite{deng2022depth, roessle2022dense} with on-the-fly single-image depth estimation. However, learning precise image-to-depth mappings is challenging due to domain gaps. A model weakly capturing such relations may produce vague depth maps or even cannot identify near and far fields.



An intuitive solution is to learn from large-scale data or employ side information such as normal \cite{Ranftl2021, Ranftl2020, yin2021learning}, or pretrained knowledge as guidance \cite{wu2022toward}. However, they require extra information burdens. Without those resources when training on data of limited appearance and depth variation, referring to \textit{scene variety}, with an extreme case that only sparse and irrelevant RGB-D pairs are available, networks can barely learn a valid image-to-depth mapping (Fig.~\ref{sv}).

Inspired by meta-learning's advantages of domain generalizability, training robust generalization models to achieve better results on unknown domains, usually learned from limited-source data \cite{finn2017model, finn2019online, nichol2018first, hospedales2021meta}, we pioneer to dig into how meta-learning applies to single-image depth prediction.
The commonly-used meta-learning problem setup follows the context of few-shot multitask settings, where a task represents a distribution to sample data from, and most tasks are designed for image classification \cite{hospedales2021meta}. Unlike those works, we study a more complex problem of scene depth estimation: the difficulties lie in per-pixel and continuous range values as outputs, in contrast to global and discrete outputs for image classification. Even for the same environments, images and depth captures can vary greatly, such as adjacent frames for a close-view object can be large room spaces. This observation indicates that our tasks are without clear task boundaries under meta-learning's context~\cite{he2019task}, and thus we propose to treat each training sample as a \textbf{fine-grained task}.


We follow the gradient-based meta-learning, which adopts a meta-optimizer and a base-optimizer \cite{finn2017model, nichol2018first}.
The base-optimizer explores multiple inner steps to find weight-updating directions. 
Then the meta-optimizer updates the meta-parameters following the explored trends. 
After few epochs of bilevel training, we learn a mapping function $\theta^{prior}$ from image to depth. It becomes better initialization for the subsequent supervised learning (Fig.~\ref{meta-inidepth}). We explain the improvements lie in progressive learning style.
Note that meta-learning and the following supervised learning operate on the same training set without using extra data.  

Previous study \cite{wu2022toward} points out indoor depth is especially hard to robustly resolve from 2D due to \textit{complex object arrangements in near fields} and \textit{arbitrary camera poses} that capture scenes from wide ranges of viewpoints. It contrasts driving scenes \cite{Dijk_2019_ICCV, watson2021temporal, tonioni2019real, watson2019self}, where decent scene a priori is given by that sky and roads dominate the upper and lower parts, and the yaw angle takes up an anchored camera rotation.
Furthermore, in addition to complex scene composition and object arrangement in the near fields, surface textures or decorations can cause confusion. A depth estimator needs to separate \textit{depth-relevant/-irrelevant low-level cues}. The former indicates depth changes along with color and appearance, such as boundaries between objects and background; the latter is surface textures where depth is invariant to colors such as material patterns of walls or paintings \cite{Chen2021S2RDepthNet,wu2022toward,wu2019salient}.

We show that meta-learning induces a prior with \textbf{higher generalizability to unseen scenes with better image-to-depth understanding}, which can identify depth-relevant/-irrelevant cues more robustly and suppress depth-irrelevant cues.
To validate the generalizability brought by meta-learning, we adopt multiple popular indoor datasets \cite{roberts2020hypersim, straub2019replica,ramakrishnan2021habitat,wu2022toward} and devise protocols for \textbf{zero-shot cross-dataset evaluation}.
This greatly differs from most previous works focusing only on intra-dataset evaluation, training and testing on a single dataset, which does not validate in-the-wild performances for practical use, such as applying to user-collected data by different cameras. 
We qualitatively and quantitatively show consistently superior performance by meta-learning on various network structures, including general and dedicated depth estimation architecture.
The work not only focuses on improvements in depth estimation. From meta-learning's perspective, we introduce fine-grained task on a continuous, per-pixel, and real-valued regression problem to advance meta-learning's study on practical problems.

\noindent\textbf{Contributions:}
\begin{itemize}[leftmargin=*,topsep=-6pt,itemsep=-0.6ex]
  \item The first method to apply meta-learning on pure single image depth prediction to achieve more generalizable and higher-performing image-to-depth understanding without using additional training data, side information, or pretrained networks.
  \item A novel fine-grained task concept in meta-learning to overcome the challenging single-image setting without obvious task boundaries. This becomes an empirical study for a complicated and practical target in meta-learning.
  \item A devised protocol for zero-shot cross-dataset evaluation of indoor scenes to faithfully evaluate a model's robustness and generalizability. Extensive experiments validate effectiveness of our meta-initialization strategy to learn a better image-to-depth prior.
  
  
\end{itemize}

%% file: 02_related.tex
\begin{figure*}[bt!]
    \centering
    \includegraphics[width=0.75\linewidth]{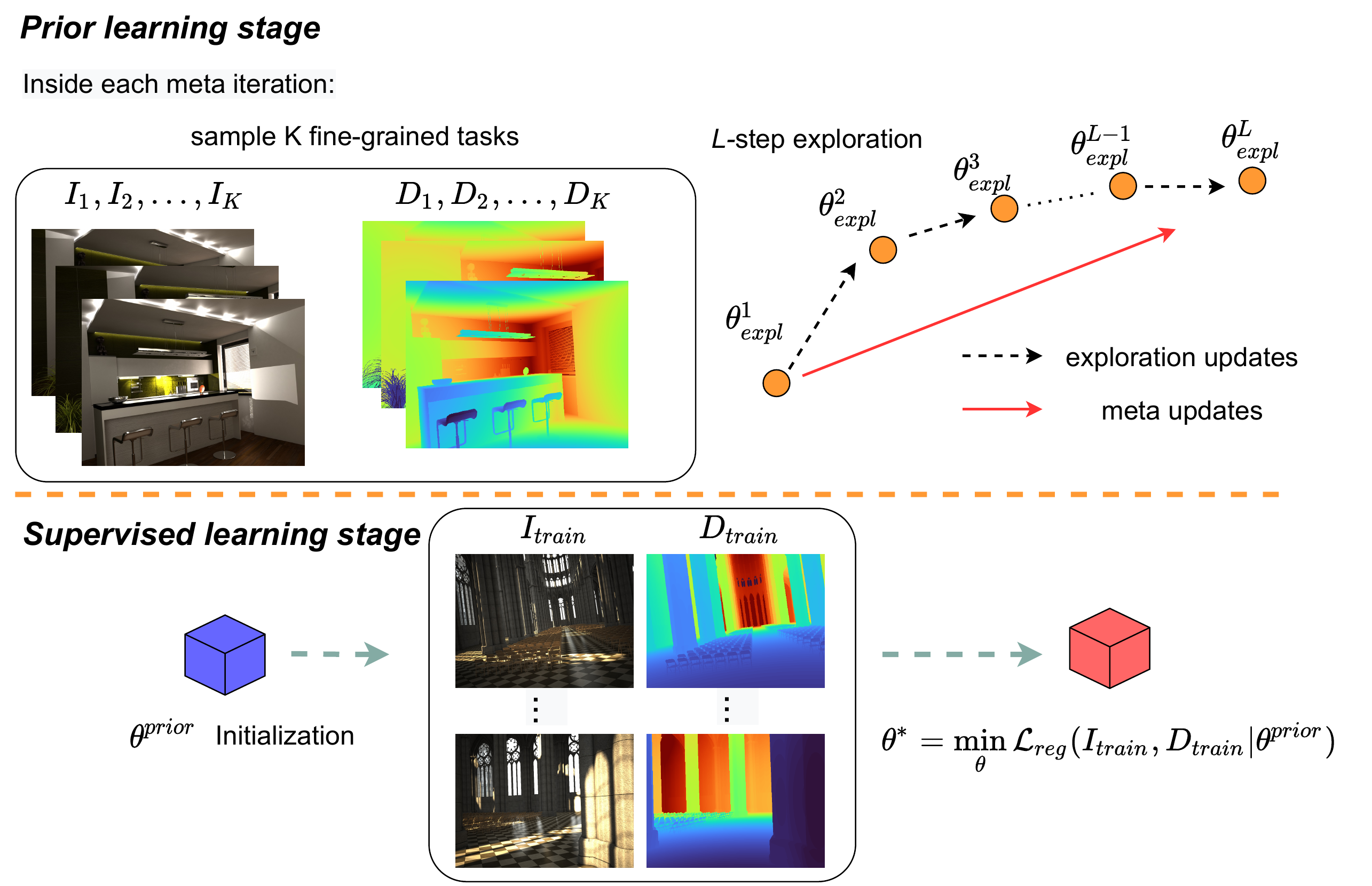}
    \vspace{-5pt}
    \caption{\textbf{Meta-Initialization for learning image-to-depth mappings.} The prior learning stage adopts a base-optimizer and a meta-optimizer. Inside each meta-iteration, $K$ fine-grained tasks are sampled and used to minimize regression loss. $L$ steps are taken by the base-optimizer to search for weight update directions for these $K$ tasks. Then, the meta-optimizer follows the explored inner trends to update meta-parameters in the Reptile style \cite{nichol2018first}. Image-to-depth prior $\theta^{prior}$ is output at the end of the stage. $\theta^{prior}$ is then used as the initialization for the subsequent supervised learning for the final model $\theta^*$.}
    \vspace{-12pt}
    \label{meta-inidepth}
\end{figure*}

\section{Related Work}
\label{related}
\subsection{Depth from Single Indoor Images}
Estimating depth from single images for indoor scenes \cite{Occlusion2018,yu2020p,zhou2019moving,ji2021monoindoor,li2022depthformer,li2021structdepth,bian2021auto,jiang2021plnet,jun2022depth,ramamonjisoa2019sharpnet,NEURIPS2019_e2c61965,wu2023inspacetype} has gained higher popularity with more high-quality datasets become available, such as Hypersim \cite{roberts2020hypersim}, Replica \cite{straub2019replica}, HM3D \cite{ramakrishnan2021habitat}, and VA \cite{wu2022toward} for photorealistic renderings from 3D environments.
In addition to fundamental pixel-wise regression loss in supervised learning, some methods adopt normal \cite{yin2019enforcing, bae2022irondepth}, plane constraints \cite{P3Depth,lee2019big,li2021structdepth,yu2020p,jiang2021plnet}, advanced loss functions \cite{fu2018deep, bhat2021adabins,liuva}, auxiliary depth completion \cite{guizilini2021sparse}, mixed-dataset training \cite{yin2021learning, Ranftl2020,Ranftl2021, zoedepth}, or module customized for depth estimation \cite{yuan2022new,kim2022global,bhat2022localbins,li2022depthformer,li2022binsformer}. In contrast, our work focuses on designing a better learning scheme: using meta-learning to improve generalizability for better image-to-depth understanding with the simplest $L_2$ regression loss and off-the-shelf architecture, which makes our analysis clear and without loss of generalization in methodology.

Most indoor-focused works are limited in intra-dataset evaluation, mainly on NYUv2 \cite{silberman2012indoor} with outdated camera models and almost frontal camera pose that cannot verify a model's robustness for practical use. We instead devise zero-shot cross-dataset evaluation protocols using several recent high-quality synthetic and real datasets. 


\subsection{Gradient-based Meta-Learning}
\vspace{-6pt}
Meta-Learning principles \cite{hochreiter2001learning,schmidhuber1987evolutionary,wu2019nonconvex} illustrate an oracle for learning how to learn, especially for domain adaption, generalization, and few-shot learning purposes. Popular gradient-based algorithms such as MAML \cite{finn2017model} and Reptile \cite{nichol2018first} are formulated as bilevel optimization with a base- and meta-optimizer. MAML uses gradients computed on the query set to update the meta-parameters. Reptile does not distinguish support and query set and simply samples data from task distribution and updates meta-parameters by differences between inner and meta-parameters. We refer readers to \cite{hospedales2021meta} for algorithm survey \cite{andrychowicz2016learning, finn2017model,rajeswaran2019meta,nichol2018first,chua2021fine,antoniou2019train,yin2020meta,rajendran2020meta,ni2021close,yao2021improving}. 

The majority of meta-learning for vision focuses on image classification \cite{balaji2018metareg,shu2021open,bui2021exploiting,qiao2020learning,dou2019domain,yao2021improving,lee2019meta,li2018learning,jamal2019task,chen2021meta,zhao2021learning,bai2021person30k,wu2019efficient} or pixel-level classification \cite{luo2022towards,lee2022pixel,cao2019meta,amac2022masksplit,gong2021cluster,guo2021metacorrection,tian2020differentiable}. \cite{gao2022matters} pioneers to study on image regression. Still, their studied problem is considered simple that regresses a rotation angle for single synthetic object. Objects are usually overlaid on all-white or out-of-context images which are far from real. 




Some works use meta-learning for depth but only for driving scenes with very different problem setup.
~\cite{zhang2019online,tonioni2019learning,zhang2020online} are built under online learning and adaptation using stereo or monocular videos for temporal consistency. They require affinity in nearby frames and meta-optimize within a single sequence, while our method is non-online, purely for single images, and meta-optimizes across multiple environments in a training set without temporal/ stereo frames. \cite{sun2022learn} groups several driving datasets as tasks but requires multiple training sets. We are the first to study meta-learning for depth from single images without assuming nearby frame affinity. The problem is arguably harder due to high appearance variation for single images from various environments. We propose fine-grained task to fulfill learning from pure single images.


%% file: 03_methods.tex
\section{Methods}
\label{methods}

\begin{figure*}[bt!]
    \centering
    \includegraphics[width=0.97\linewidth]{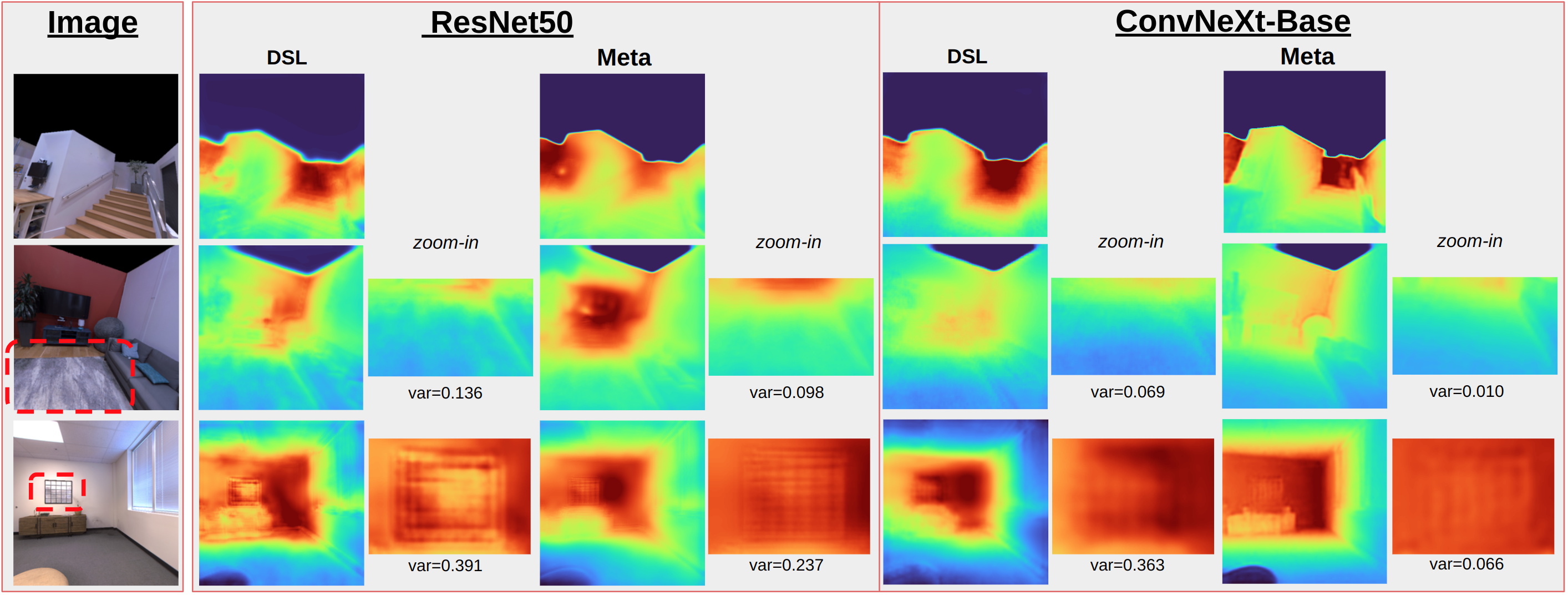}
    \vspace{-4pt}
    \caption{\textbf{Fitting to training environments.} \textit{var} shows depth variance in the highlighted regions. We show comparison of fitting to training environments between pure meta-learning (Meta) and direct supervised learning (DSL) on limited scene-variety dataset, Replica. Meta produces smooth and more precise depth. Depth-irrelevant textures on planar regions can be resolved more correctly. In contrast, DSL produces irregularities affected by local high-frequency details, especially ResNet50. See Sec. \ref{experiments:scene-fitting} for details and \ref{explanation} for the explanation. 
    }
    \label{fitting}
    \vspace{-13pt}
\end{figure*}

\subsection{Difficulty in Accurate Depth Estimation}
A model needs to distinguish \textit{depth-relevant} and \textit{depth-irrelevant low-level cues} to accurately estimate depth from images. The former shows color or radiance changes at object boundaries to the background. In the latter case, geometry is invariant to color changes, such as decoration or object textures (see Supp Sec. C). Depth-irrelevant cues frequently appear in indoor scenes due to cluttered textured objects in near fields.
For example, a painting triggers many depth-irrelevant features and confuses networks. It then heavily relies on sufficient variety 
in training data, which demonstrate mappings from images to depth and enable learning from global context to suppress such local high-frequency details.
For example, \cite{Ranftl2021, Ranftl2020, zoedepth} train on mixed data sources to robustly estimate depth in the wild. When training set size is limited, there are no sufficient examples to describe mappings between the two domains. A model can either not be able to estimate precise depth or simply memorize seen image-depth pairs but cannot generalize to unseen examples \cite{arpit2017closer,feldman2020neural}. See Fig.~\ref{sv}. Thus, we exploit meta-learning with its advantages that without extra data sources, it attains few-shot learning and higher generalizability. To adapt meta-learning to single-image depth estimation, we propose fine-grained task as follows. 


\subsection{Single RGB-D Pair as Fine-Grained Task}
\vspace{-6pt}

\textbf{Definition}.
Single-image depth prediction learns a function $f_{\theta}:\mathcal{I}\to\mathcal{D}$, parameterized by $\theta$, to map from imagery to depth. A training set ($\mathbb{I}_{train}$, $\mathbb{D}_{train}$), containing image $I \in \mathbb{I}_{train}$ and associated depth map $D \in \mathbb{D}_{train}$, is used to train a model. In a minibatch with size $K$, each pair ($I_i$, $D_i$), $\forall i \in [1, K]$ is treated as a \textbf{fine-grained task}. Fine-grained tasks are mutual-exclusive: no two scenes sampled from the meta-distribution, i.e., the whole RGB-D dataset, share the same scene appearance and depth relation. Proof: assume we have two different scene images $I_1$ and $I_2$, and each contains sets of regions $\mathbb{R}_1$ and $\mathbb{R}_2$. The null set $\phi \notin \mathbb{R}^-$, where $\mathbb{R}^- = (\mathbb{R}_1 - \mathbb{R}_2) \cup (\mathbb{R}_2 - \mathbb{R}_1)$ that contains regions appear only in either $I_1$ or $I_2$, since $I_1$ and $I_2$ are different frames and inevitably capture different regions. Thus, any two scenes have different appearance and depth relations.

\textbf{Difference with task in meta-learning}.
Fine-grained tasks are different from tasks in most-used meta-learning or few-shot learning usages \cite{finn2017model}, where a task contains data distribution and batches are sampled from it. 
Fine-grained tasks do not contain data distribution but are sampled from meta-distribution, the whole RGB-D dataset.  
For example, a navigating agent captures image and depth pairs. The RGB-D pairs are sampled from the meta-distribution.

\textbf{Design}. Each fine-grained task is used to learn on the specific RGB-D pair. 
The design is motivated by the fact that appearance and depth variation can be high. 
A view looking at small desk objects and a view of large room spaces are highly dissimilar in contents and ranges.
Mappings from their scene appearance to range values are different. Still, they can be captured in the same environment or even in neighboring frames.
This contrasts with image classification where class samples share a common label. 
The observation explains why we treat each RGB-D pair as a fine-grained task instead of each environment.


\subsection{Meta-Initialization on Depth from Single Image}
We describe our approach based on gradient-based meta-learning to learn a good initialization (Fig. \ref{meta-inidepth}).

\textbf{Prior learning stage}.
 In the first prior learning stage, we adopt a meta-optimizer and a base-optimizer. In each meta-iteration, $K$ fine-grained tasks as a minibatch are sampled from the whole training set: $(I_i, D_i)\sim (\mathbb{I}_{train}, \mathbb{D}_{train})$, $\forall i \in [1,K]$. Then we take $L$ steps to explore gradient directions that minimize the regression loss and get ($\theta^1_{expl}, \theta^2_{expl}, ..., \theta^L_{expl}$):
\begin{equation}
    \vspace{-6pt}
      \NORMAL{\theta^{i}_{expl} \leftarrow \theta^{i-1}_{expl} - \alpha  \frac{1}{K}  \nabla_{\theta}\sum_{k\in [1,K]} \mathcal{L}_{reg}(I_k, D_k; \theta^{i-1}_{expl}), \forall i \in [1,L].}
     \vspace{-6pt}
\label{exploration}
\end{equation}
After the $L$-step exploration, we update the meta-parameters using Reptile style\cite{nichol2018first}, i.e., following the explored weight updating direction in the inner steps. 
\begin{flalign}
    \vspace{-2pt}
     \qquad \qquad \theta^{j}_{meta} \leftarrow \theta^{j-1}_{meta} - \beta (\theta^{j-1}_{meta} - \theta^L_{expl}), &&
     \vspace{-2pt}
\label{meta-update}
\end{flalign}
where $\alpha$ and $\beta$ are respective learning rates. $i$ and $j$ denote inner and meta-iterations.

\begin{figure}[bt!]
    \centering
    \includegraphics[width=1.00\linewidth]{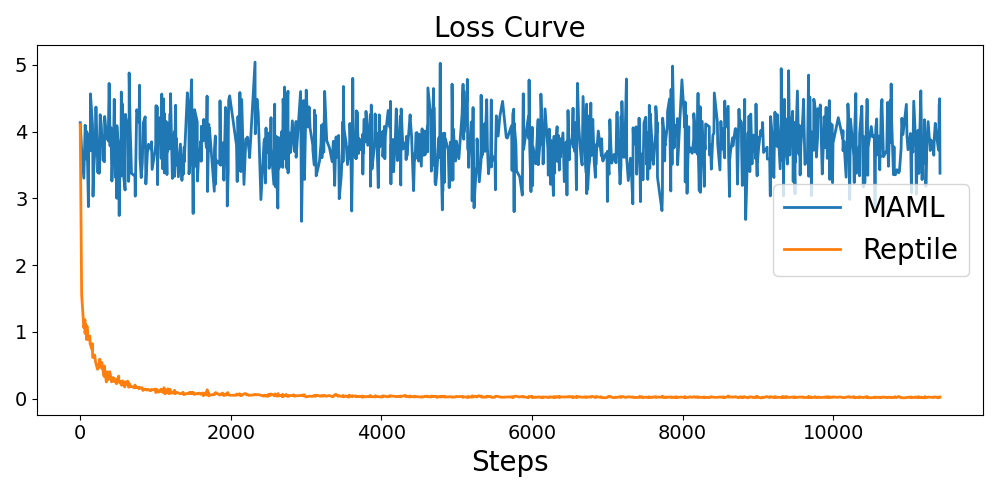}
    \vspace{-18pt}
    \caption{\textbf{Loss curve for MAML v.s. Reptile.}
    }
    \label{loss_curve}
    \vspace{-12pt}
\end{figure}

Compared with MAML \cite{finn2017model}, we find Reptile is more suitable for training for fine-grained tasks. First, as mentioned in Reptile's paper \cite{nichol2018first}, it is designed without support and query set split, and thus it inherently does not require multiple data samples in a task, which matches our fine-grained task definition. Next, first-order MAML computes gradients on the query set at the last inner step $\theta^L_{expl}$ to update meta-parameters. However, only one sample exists in each fine-grained task, and each fine-grained task is mutual-exclusive and can differ greatly, depending on $\mathbb{R}^-$. Thus, if taking exploration on a support split and computing gradients on the query split, but the support and query samples do not share common components, the gradients are nearly random and prevent from converging. By contrast, Reptile does not entail support and query split or require common components between samples, so it stabilizes training towards convergence and becomes the choice. We show the loss curve in Fig. \ref{loss_curve}.

\textbf{Supervised learning stage.}
Prior knowledge $\theta^{prior}$ is learned after the first stage. We treat it as the initialization for the subsequent supervised learning with conventional stochastic gradient descent to minimize the regression loss.
\begin{equation}
\vspace{-2pt}
     \theta^* \leftarrow \min_{\theta} \mathcal{L}_{reg}(\mathbb{I}_{train},\mathbb{D}_{train}|\theta^{prior}).
     \vspace{-2pt}
\label{supervised}
\end{equation}
Last, the test set ($\mathbb{I}_{test}, \mathbb{D}_{test}$) is used to evaluate the depth estimation performance of $\theta^*$.
Algorithm \ref{meta-algo} organizes the whole procedure, and we show the pseudo-code in Supp Sec. G. The implementation only needs a few-line codes as plugins to depth estimation frameworks, which benefits higher model generalizability as shown in later experiments. 

\textbf{Difference with other learning strategies.}
Compared with widely-used pretraining that requires multiple data sources to gain generalizability \cite{Ranftl2020, yin2021learning, Ranftl2021}, both the prior learning and supervised learning stages operate on the same dataset without access to extra data or off-the-shelf models. Thus, they are free from those burdens. 

Compared with simple gradient accumulation \cite{ruder2016overview}, where gradients are accumulated for several batches and then used to update parameters only once, the bilevel optimization keeps updating the inner-parameters every step in $L$ to find the local niche for the current batch. Besides, gradient accumulation has the effect of large batch size, which might cause overfitting and degrade model generalizability.

\begin{algorithm}
\caption{Our Meta-Initialization Procedure}
\begin{algorithmic}[1]
    \FOR {epoch = $1:N$}
    	\FOR {$j = 1:T$ (iterations)}
    		\STATE $\theta^0_{expl}$ $\leftarrow$ $\theta^j_{meta};\ $ ($I_1$, $D_1$), ($I_2$, $D_2$) $...$ ($I_K$, $D_K$) $\sim$ ($\mathbb{I}_{train}$, $\mathbb{D}_{train}$). 
    		\FOR {$i$ = $1:L$ (steps)}
    		    \scriptsize \STATE $\theta^{i}_{expl} \leftarrow \theta^{i-1}_{expl} - \alpha  \frac{1}{K} \nabla_{\theta}\sum_{k\in [1,K]} \mathcal{L}_{reg}(I_k, D_k; \theta^{i-1}_{expl})$.
    		\ENDFOR
    		\STATE $\theta^{j}_{meta} \leftarrow \theta^{j-1}_{meta} - \beta (\theta^{j-1} - \theta^L_{expl})$.
		\ENDFOR
    \ENDFOR
    \STATE Prior $\theta^{prior} \leftarrow \theta_{meta}^T$ at epoch $N$.
    \STATE Use $\theta^{prior}$ as initialization. Supervised learning by $\theta^* \leftarrow \min_{\theta} \mathcal{L}_{reg}(\mathbb{I}_{train},\mathbb{D}_{train}|\theta^{prior})$.
\end{algorithmic}
\label{meta-algo}
\vspace{-4pt}
\end{algorithm}

\subsection{Strategy and Explanation}
\label{explanation}
\textbf{Meta-Initialization.}
We next analyze meta-learning behavior with fine-grained task.
Inside each meta-iteration, the base-learner explores the neighborhood with $L$-step using $K$ fine-grained tasks. Compared with simple single-step update, the meta-update can be seen as first taking $L$-step amortized gradient descent with a lower learning rate to delicately explore local loss manifolds, then updating meta-parameters by trends shown in the inner steps with a step size $\beta$ towards $\theta^L$.
$\theta^{prior}$ after the prior learning may underfit the training set since the algorithm suggests not wholly following optimal gradients for each batch but with a $\beta$ for control. However, it avoids overfitting to seen RGB-D pairs and forces the inner exploration to reach higher-level image-to-depth understanding. $\theta^{prior}$ then becomes good initialization for downstream RGB-D learning.

\begin{figure}[bt!]
    \centering
    \includegraphics[width=0.92\linewidth]{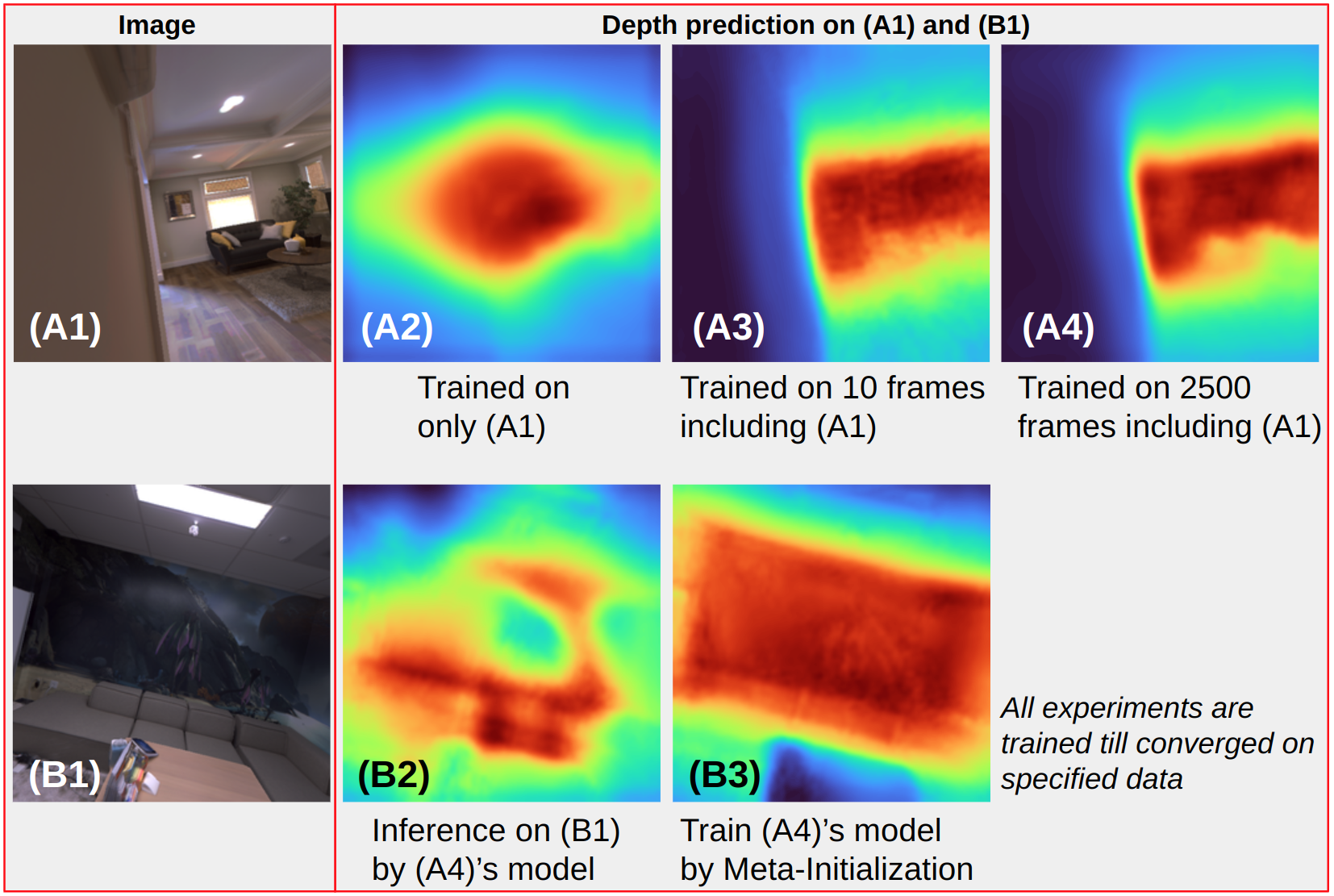}
    \vspace{-3pt}
    \caption{\textbf{Analysis on scene variety and model generalizability.} (A) shows limited training scenes constrain learning image-to-depth mappings, with an extreme case (A2) for only one training image. (B) shows though a model (A4) fits well on training scenes, it still cannot generalize to unseen seen, especially wall paintings with many depth-irrelevant cues. Meta-initialization attains better model generalizability. See Sec. \ref{explanation} for explanation.
    }
    \label{sv}
    \vspace{-15pt}
\end{figure}

\textbf{Progressive learning perspective.}
Algorithm \ref{meta-algo} can be seen as progressive learning on a training set. At the first stage, meta-learning benefits learning coarse but smooth depth from global context. In Fig. \ref{fitting} we apply the first-stage meta-learning compared to direct supervised learning on a dataset of limited scene variety. Meta-learning estimates smooth depth shapes and is free from irregularity that direct supervised learning encounters. 
The irregularity indicates the dataset did not provide sufficient scene variety that demonstrates how images map to depth in various environments to learn smooth depth from global context. Consequently, only local high-frequency cues show. To illustrate, if only sparse and irrelevant scene images are presented, finding a function that satisfactorily fits those scenes with smooth depth from global context is hard. See Fig.~\ref{sv}.
The irregularity occurs especially at cluttered objects or surface textured areas, since those \textit{depth-irrelevant local cues are barely suppressed}. 
In summary, the progressive fashion first learns coarse but smooth depth by $\theta^{prior}$. Then, the network learns finer depth at the second supervised stage.

%% file: 04_exp.tex
\section{Experiments and Discussion}
\label{experiments}
\begin{table*}[h]
\begin{center}
  \caption{\textbf{Generalizability with different scene variety.} We compare single-stage meta-learning (only prior learning) and supervised learning. ConvNeXt-Base backbone is used. $a \to b$ means training on $a$- and testing on $b$-dataset. Replica and HM3D respectively hold lower and higher scene variety for training. 
   Meta-Learning has much larger improvements especially trained on low scene-variety Replica.}
  \vspace{-1pt}
  \footnotesize
  \label{table:limited}
  \begin{tabular}[c]
  {
  p{3.3cm}<{\arraybackslash}|
  p{1.4cm}<{\centering\arraybackslash}|
  p{1.4cm}<{\centering\arraybackslash}|
  p{1.4cm}<{\centering\arraybackslash}|
  p{1.4cm}<{\centering\arraybackslash}|
  p{1.4cm}<{\centering\arraybackslash}|
  p{1.4cm}<{\centering\arraybackslash}}
  \hlineB{2}
  \hline
     & \multicolumn{3}{c}{\cellcolor[HTML]{99CCFF}Replica $\to$ VA} & \multicolumn{3}{c}{\cellcolor[HTML]{F7BCDC}HM3D $\to$ VA} \\
      Method & \cellcolor[HTML]{FAE5D3} MAE & \cellcolor[HTML]{FAE5D3} AbsRel & \cellcolor[HTML]{FAE5D3} RMSE & \cellcolor[HTML]{FAE5D3} MAE & \cellcolor[HTML]{FAE5D3} AbsRel & \cellcolor[HTML]{FAE5D3} RMSE\\
    \hline
      Direct supervised learning & 0.718 & 0.538 & 1.078 & 0.544 & 0.456 & 0.715 \\
      Meta-Learning & \textbf{0.548} & \textbf{0.430} & \textbf{0.761} & \textbf{0.427} & \textbf{0.369} & \textbf{0.603} \\
      Improvement & \textit{-23.6\%} & \textit{-20.1\%} & \textit{-29.4\%} & \textit{-21.5\%} & \textit{-19.1\%} & \textit{-15.7\%} \\
    \hline
    \hlineB{2}
  \end{tabular}
  \vspace{-8pt}
\end{center}
\end{table*}

\textbf{Aims}. We validate our meta-initialization with five questions. \textbf{Q1} Can meta-learning improve learning image-to-depth mapping on limited scene-variety datasets? (Sec.~\ref{experiments:scene-fitting}) \textbf{Q2} What improvements can meta-initialization bring compared with the most popular ImageNet-initialization? (Sec.~\ref{experiments:intra-dataset}) \textbf{Q3} How does meta-initialization help zero-shot cross-dataset generalization? (Sec.~\ref{experiments:cross-dataset}) \textbf{Q4} How does more accurate depth help learn better 3D representation? (Sec.~\ref{experiments:nerf}) \textbf{Q5} How is the proposed fine-grained task related to other meta-learning findings? (Supp Sec. D)


\textbf{Datasets:} Exemplar data are illustrated in Supp Sec. B.
\begin{itemize}[leftmargin=*,topsep=-3pt,itemsep=-0.8ex]
  \item Hypersim \cite{roberts2020hypersim} has high scene variety with 470 synthetic indoor environments, from small rooms to large open spaces, with about 67K training and 7.7K testing images.
  \item HM3D \cite{ramakrishnan2021habitat} and Replica \cite{straub2019replica} are associated with 200K and 40K images that are taken from SimSIN \cite{wu2022toward}. HM3D has 800 scenes with \textit{much higher scene variety} than Replica, which only has 18 overlapping scenes. 
  \item NYUv2 \cite{silberman2012indoor} contains real 654 testing images. It uses older camera models with high imaging noise and limited camera viewing direction.
  \item VA \cite{wu2022toward} consists of 3.5K photorealistic renderings for testing on challenging lighting conditions and arbitrary camera viewing directions.
\end{itemize}
\vspace{4pt}

\textbf{Training Settings.} We use ResNet \cite{he2016deep}, ConvNeXt \cite{liu2022convnet}, and their variants as the network architecture to extract bottleneck features. Then we build a depth regression head following \cite{Godard_2019_ICCV} containing 5 convolution blocks with skip connection from the encoder. Each convolution block contains a 3x3 convolution, an ELU activation, and a bilinear 2$\times$ upsampling layer to recover the input size at the end output. Channel-size of each convolution block is $(256, 128, 64, 32, 16)$. Last, a 3x3 convolution for 1-channel output with a sigmoid activation are used to get inverse depth, which is then converted to depth \cite{Dijk_2019_ICCV}. We set $N=5$, $L=4$, $K=50$, $(\alpha, \beta)=(0.001, 0.5)$ for ResNet, and $(\alpha, \beta)=(0.0005, 0.5)$ for ConvNeXt. At the supervised learning stage, we train models for 15 epochs with a learning rate of $3\times10^{-4}$, optimized by AdamW \cite{loshchilov2018decoupled} with a weight decay of 0.01. Input size to the network is 256$\times$256.  $L_2$ loss is used as $\mathcal{L}_{reg}$. 



\textbf{Metrics.} We adopt common monocular depth estimation evaluation metrics. Error metrics (in meters, the lower the better): Mean Absolute Error (MAE), Absolute Relative Error (AbsRel), Root Mean Square Error (RMSE). Threshold Accuracy: $\delta_C$ (in $\%$, percentage of correct pixel depth. Higher percentage implies more structured depth hence better). Correctness: ratio between prediction and groundtruth is within $1.25^C, C=[1,2,3]$.

\subsection{Meta-Learning on Limited Scene Variety} 
\label{experiments:scene-fitting}
We first show how a single-stage meta-learning (only prior stage) performs. We train $N$=15 epochs of meta-learning and compare with 15 epochs of direct supervised learning where both training pieces converge already. 
The other hyperparameters are the same as given in Training Settings. 
Replica Dataset of limited scene variety is used to verify gain on limited sources.
Fig. \ref{fitting} show fitting to training scenes. 
From the figure, meta-learning is capable of identifying near and far fields without irregularity that direct supervised learning struggles with. Under limited training scenes, meta-learning induces a better image-to-depth mapping that delineates object shapes, separates depth-relevant/-irrelevant cues, and shows flat planes where rich depth-irrelevant textures exist. The observation follows the explanation in Sec.~\ref{explanation}. See Supp Sec. C for more analysis.

We next numerically examine generalizability to unseen scenes when training on different level scene-variety data. HM3D (high scene variety) and Replica (low scene variety) are used as training sets and VA is used for testing. Table \ref{table:limited} shows that models trained by single-stage meta-learning substantially outperform direct supervised learning with $15.7\%$-$29.4\%$ improvements. The advantage is more evident when trained on lower scene-variety Replica. 

\begin{table}[tb!]
\label{table:intra}
\begin{center}
  \caption{\textbf{Effects of Meta-Initialization on intra-dataset evaluation.} We train and test meta-initialization (full Algorithm \ref{meta-algo}) on the same dataset. Hypersim and NYUv2 of higher scene variety are used. Using the same architecture, meta-initialization (+Meta) consistently outperforms ImageNet-initialization (no marks). Both error (in orange) and accuracy (in green) are reported.}
  \vspace{-8pt}
  \footnotesize
  \label{table:intra-dataset}
  \begin{tabular}[c]
  {
  p{2.7cm}<{\arraybackslash}
  p{0.5cm}<{\centering\arraybackslash}
  p{0.65cm}<{\centering\arraybackslash}
  p{0.6cm}<{\centering\arraybackslash}
  p{0.4cm}<{\centering\arraybackslash}
  p{0.4cm}<{\centering\arraybackslash}
  p{0.4cm}<{\centering\arraybackslash}}
  \hlineB{2}
  
      \multicolumn{1}{|c|}{\cellcolor[HTML]{99CCFF} Hypersim}  & \cellcolor[HTML]{FAE5D3} MAE & \cellcolor[HTML]{FAE5D3} AbsRel & \cellcolor[HTML]{FAE5D3} RMSE &  \multicolumn{1}{|c|}{\cellcolor[HTML]{D5F5E3} $\delta_1$} &  \multicolumn{1}{|c|}{\cellcolor[HTML]{D5F5E3} $\delta_2$} &  \multicolumn{1}{|c|}{\cellcolor[HTML]{D5F5E3} $\delta_3$} \\
    \hline
      ResNet50     & 1.288 & 0.248 & 1.775 & 64.8 & 87.1 & 94.7 \\
      ResNet50 + Meta & \textbf{1.205} & \textbf{0.239} & \textbf{1.680} & \textbf{66.7} & \textbf{87.9} & \textbf{95.0} \\
      \hline
      ResNet101    & 1.197 & 0.234 & 1.671 & 67.4 & 88.5 & 95.3 \\
      ResNet101 + Meta   & \textbf{1.158} & \textbf{0.220} & \textbf{1.595} & \textbf{68.0} & \textbf{89.0} & \textbf{95.4}\\
      \hline
      ConvNeXt-base   & 1.073 & 0.201 & 1.534 & 73.6 & 91.1 & 96.3 \\
      ConvNeXt-base + Meta  & \textbf{0.994} & \textbf{0.188} & \textbf{1.425} & \textbf{74.9} & \textbf{91.7} & \textbf{96.5} \\ 
    \hlineB{2}
    \hline
  \end{tabular}

  \begin{tabular}[c]
  {
  p{2.7cm}<{\arraybackslash}
  p{0.5cm}<{\centering\arraybackslash}
  p{0.65cm}<{\centering\arraybackslash}
  p{0.6cm}<{\centering\arraybackslash}
  p{0.4cm}<{\centering\arraybackslash}
  p{0.4cm}<{\centering\arraybackslash}
  p{0.4cm}<{\centering\arraybackslash}}
  \hlineB{2}
  
     \multicolumn{1}{|c|}{\cellcolor[HTML]{99CCFF} NYUv2}  & \cellcolor[HTML]{FAE5D3} MAE & \cellcolor[HTML]{FAE5D3} AbsRel & \cellcolor[HTML]{FAE5D3} RMSE &  \multicolumn{1}{|c|}{\cellcolor[HTML]{D5F5E3} $\delta_1$} &  \multicolumn{1}{|c|}{\cellcolor[HTML]{D5F5E3} $\delta_2$} &  \multicolumn{1}{|c|}{\cellcolor[HTML]{D5F5E3} $\delta_3$} \\
    \hline
      ResNet50    & 0.345 & 0.131 & 0.480 & 83.6 & 96.4 & 99.0 \\
      ResNet50 +Meta   & \textbf{0.325} & \textbf{0.122} & \textbf{0.454} & \textbf{85.4} & \textbf{96.8} & \textbf{99.3} \\
      \hline
      ResNet101    & 0.318 & 0.120 & 0.448 & 85.6 & 97.1 & 99.3 \\
      ResNet101 +Meta   & \textbf{0.303} & \textbf{0.112} & \textbf{0.420} & \textbf{86.7} & \textbf{97.4} & \textbf{99.4} \\
      \hline
      ConvNeXt-base   & 0.273 & 0.101 & 0.394 & 89.4 & 97.9 & \textbf{99.5} \\ 
      ConvNeXt-base+Meta   & \textbf{0.266} & \textbf{0.099} & \textbf{0.387} & \textbf{89.8} & \textbf{98.1} & \textbf{99.5} \\ 
      \hline
    \hlineB{2}
    \hline
  \end{tabular}
  \vspace{-23pt}
\end{center}
\end{table}

\begin{figure*}[bt!]
    \centering
    \vspace{-5pt}
    \includegraphics[width=0.88\linewidth]{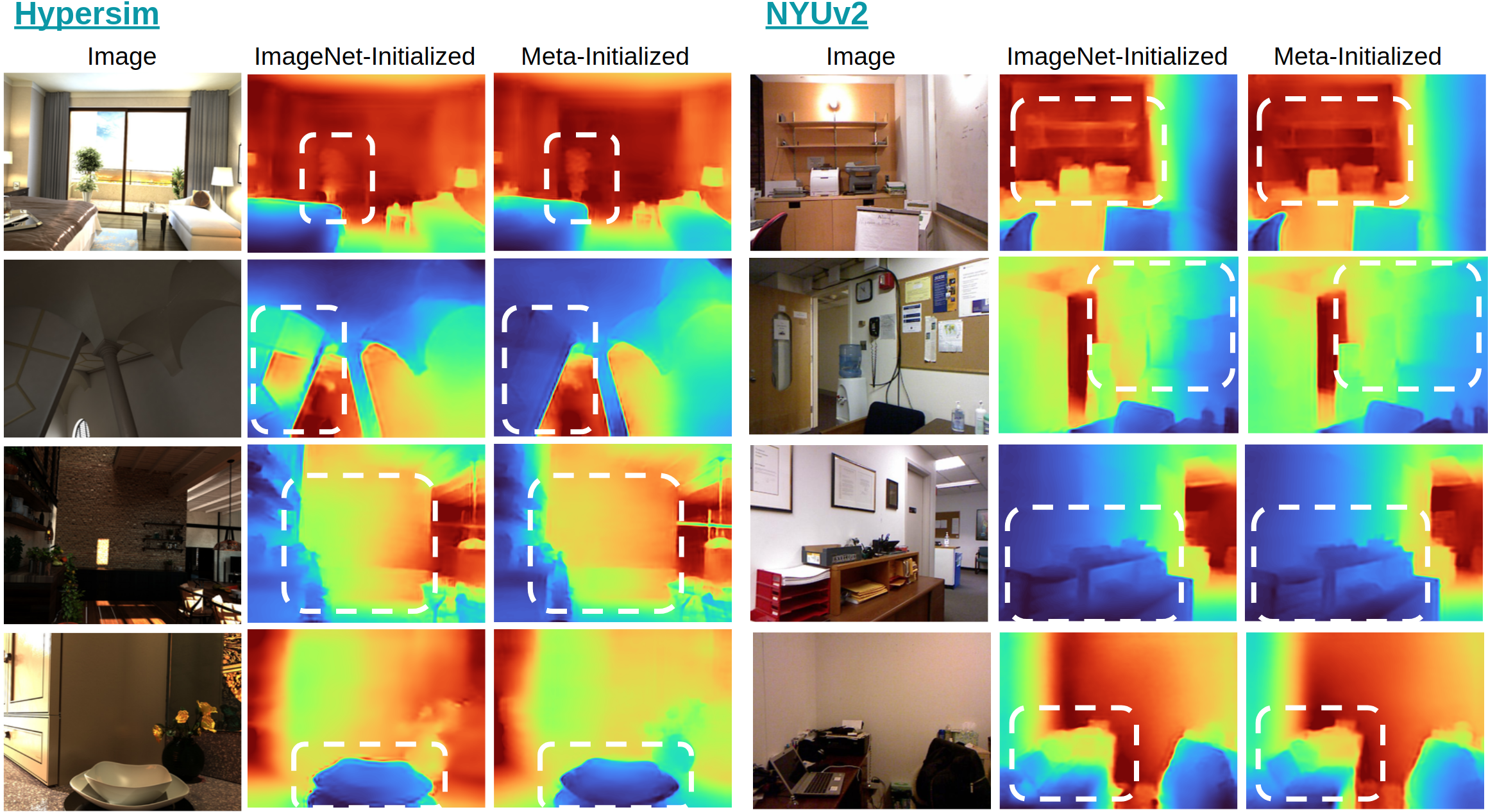}
    \vspace{-4pt}
    \caption{\textbf{Depth map qualitative comparison.} Results of our meta-initialization have better object shapes with clearer boundaries. Depth-irrelevant textures are suppressed, and flat planes are predicted, as shown in Hypersim- Row 2 ceiling and 3 textured wall examples.}
    \vspace{-10pt}
    \label{intra-qualitative}
\end{figure*}


\subsection{Meta-Initialization v.s. ImageNet-Initialization} 
\label{experiments:intra-dataset}


Sec.~\ref{experiments:scene-fitting} shows single-stage meta-learning induces much better depth regression, but the depth is yet detailed. We next train full Algorithm \ref{meta-algo}, using meta-learned weights as initialization for following supervised learning. We \textbf{go beyond limited sources} and train on higher scene-variety datasets. 
Intuitively, higher scene variety helps supervised learning attain better depth prediction and might diminish meta-learning's advantages of few-shot and low-source learning.
However, such studies are practical for validating meta-learning in real-world applications.
Comparison is drawn with baselines of direct supervised learning without meta-initialization that begins from ImageNet-initialization instead.


Table \ref{table:intra-dataset} shows intra-dataset evaluation that trains/ tests on each official data splits.
For Hypersim evaluation is capped for depth at 20m and 10m for NYUv2. 
Uniquely, using meta-initialization attains \textbf{consistently} lower errors and higher accuracy than the baselines, especially AbsRel (averagely +6.5\%) and $\delta_1$ (averagely +1.69 points) that indicates accurate depth structure is predicted. We further display qualitative comparison for depth and 3D point cloud view in Fig. \ref{intra-qualitative} and \ref{pointcloud}. 
The gain simply comes from better training schema without more data or constraints, advanced loss, or model design. More results are in Supp Sec. G.

\subsection{Zero-Shot Cross-Dataset Evaluation} 
\label{experiments:cross-dataset}

To faithfully validate a trained model in the wild, we design protocols for zero-shot cross-dataset inference. High scene-variety and larger-size synthetic datasets, Hypersim and HM3D, are used as training sets. VA, Replica, and NYUv2 serve as testing, and their evaluations are capped at 10m. We median-scale prediction to groundtruth in the protocol to compensate for different camera intrinsic. In Table \ref{table:cross-dataset-hm3d}, compared with ImageNet-initialization, meta-initialization \textbf{consistently} improves in nearly all the metrics, especially $\delta_1$ (averagely +1.97 points). 
The gain comes from that meta-prior attains a better image-to-depth mapping with coarse but smooth and reasonable depth. Conditioned on the initialization, the learning better calibrates to open-world image-to-depth relation hence generalizes better to unseen scenes. 


\begin{table}[tb!]
\begin{center}
  \caption{\textbf{Zero-Shot cross-dataset evaluation using meta-initialization (Algorithm \ref{meta-algo}).} Comparison is drawn between without meta-initialization (no marks, ImageNet-initialization) and with our meta-initialization (Meta) using different sizes ConvNeXt. Results of "+Meta" are consistently better.}
  \vspace{-4pt}
  \footnotesize
  \label{table:cross-dataset-hm3d}
  
  \begin{tabular}[c]
  {
  p{2.75cm}<{\arraybackslash}|
  p{0.5cm}<{\centering\arraybackslash}|
  p{0.65cm}<{\centering\arraybackslash}|
  p{0.6cm}<{\centering\arraybackslash}|
  p{0.4cm}<{\centering\arraybackslash}|
  p{0.4cm}<{\centering\arraybackslash}|
  p{0.4cm}<{\centering\arraybackslash}}
  \hlineB{2}
  
       \multicolumn{1}{|c|}{\cellcolor[HTML]{99CCFF} HM3D $\to$ VA} & \cellcolor[HTML]{FAE5D3} MAE & \cellcolor[HTML]{FAE5D3} AbsRel & \cellcolor[HTML]{FAE5D3} RMSE & \cellcolor[HTML]{D5F5E3} $\delta_1$ & \cellcolor[HTML]{D5F5E3} $\delta_2$ & \cellcolor[HTML]{D5F5E3} $\delta_3$ \\
    \hline
      ConvNeXt-small & 0.267 & 0.180 & 0.389 & 74.6 & 91.0 & 96.1 \\
      ConvNeXt-small + Meta & \textbf{0.233} & \textbf{0.162} & \textbf{0.345} & \textbf{77.8} & \textbf{93.1} & \textbf{97.3} \\
      ConvNeXt-base  & 0.258 & 0.176 & 0.385 & 76.1 & 91.1 & 95.4 \\ 
      ConvNeXt-base + Meta  & \textbf{0.238} & \textbf{0.163} & \textbf{0.356} & \textbf{78.0} & \textbf{92.5} & \textbf{96.7} \\
      ConvNeXt-large  & 0.242 & 0.170 & 0.357 & 78.1 & 91.5 & 95.7 \\
      ConvNeXt-large + Meta & \textbf{0.226} & \textbf{0.160} & \textbf{0.330} & \textbf{78.9} & \textbf{92.2} & \textbf{96.3} \\
    \hlineB{2}
    
  
       \multicolumn{1}{|c|}{\cellcolor[HTML]{99CCFF} HM3D $\to$ NYUv2}  & \cellcolor[HTML]{FAE5D3} MAE & \cellcolor[HTML]{FAE5D3} AbsRel & \cellcolor[HTML]{FAE5D3} RMSE & \cellcolor[HTML]{D5F5E3} $\delta_1$ & \cellcolor[HTML]{D5F5E3} $\delta_2$ & \cellcolor[HTML]{D5F5E3} $\delta_3$ \\
    \hline
      ConvNeXt-small  & 0.540 & 0.213 & 0.728 & 69.2 & 88.7 & 95.8 \\
      ConvNeXt-small + Meta & \textbf{0.527} & \textbf{0.206} & \textbf{0.710} & \textbf{70.7} & \textbf{89.0} & \textbf{95.9} \\
      ConvNeXt-base  & 0.529 & 0.208 & 0.717 & 70.1 & 89.4 & 96.0 \\
      ConvNeXt-base + Meta & \textbf{0.505} & \textbf{0.199} & \textbf{0.691} & \textbf{71.6} & \textbf{89.8} & \textbf{96.3} \\
      ConvNeXt-large & 0.501 & 0.192 & 0.690 & 72.0 & 90.4 & 96.4 \\ 
      ConvNeXt-large +Meta & \textbf{0.481} & \textbf{0.190} & \textbf{0.660} & \textbf{73.2} & \textbf{90.6} & \textbf{96.6} \\ 
    \hlineB{2}
    
    \multicolumn{1}{|c|}{\cellcolor[HTML]{99CCFF} HM3D $\to$ Replica} & \cellcolor[HTML]{FAE5D3} MAE & \cellcolor[HTML]{FAE5D3} AbsRel & \cellcolor[HTML]{FAE5D3} RMSE & \cellcolor[HTML]{D5F5E3} $\delta_1$ & \cellcolor[HTML]{D5F5E3} $\delta_2$ & \cellcolor[HTML]{D5F5E3} $\delta_3$ \\
    \hline
      ConvNeXt-small & 0.222 & 0.138 & 0.321 & 84.5 & 93.9 & 96.6 \\
      ConvNeXt-small + Meta & \textbf{0.200} & \textbf{0.126} & \textbf{0.287} & \textbf{85.6} & \textbf{95.7} & \textbf{98.1} \\
      ConvNeXt-base & 0.217 & 0.134 & 0.316 & 84.6 & 94.2 & 96.6 \\ 
      ConvNeXt-base + Meta & \textbf{0.192} & \textbf{0.117} & \textbf{0.277} & \textbf{87.1} & \textbf{96.4} & \textbf{98.5} \\
      ConvNeXt-large & 0.214 & 0.137 & 0.307 & 84.3 & 94.0 & 96.6 \\
      ConvNeXt-large + Meta  & \textbf{0.191} & \textbf{0.117} & \textbf{0.275} & \textbf{87.1} & \textbf{96.5} & \textbf{98.5} \\
    \hlineB{2}
    \hline
  
       \multicolumn{1}{|c|}{\cellcolor[HTML]{99CCFF} Hypersim $\to$ VA}  & \cellcolor[HTML]{FAE5D3} MAE & \cellcolor[HTML]{FAE5D3} AbsRel & \cellcolor[HTML]{FAE5D3} RMSE & \cellcolor[HTML]{D5F5E3} $\delta_1$ & \cellcolor[HTML]{D5F5E3} $\delta_2$ & \cellcolor[HTML]{D5F5E3} $\delta_3$ \\
    \hline
      ConvNeXt-small & 0.291 & 0.215 & 0.404 & 68.5 & 90.8 & 96.7 \\
      ConvNeXt-small + Meta & \textbf{0.280} & \textbf{0.207} & \textbf{0.398} & \textbf{70.4} & \textbf{91.3} & \textbf{97.0} \\
      ConvNeXt-base  & 0.275 & 0.201 & 0.393 & 71.3 & 91.8 & 97.3 \\
      ConvNeXt-base + Meta & \textbf{0.259} & \textbf{0.194} & \textbf{0.365} & \textbf{72.8} & \textbf{92.8} & \textbf{97.8} \\ 
      ConvNeXt-large  & 0.263 & 0.198 & 0.369 & 73.0 & 92.0 & 97.1 \\
      ConvNeXt-large + Meta  & \textbf{0.248} & \textbf{0.183} & \textbf{0.355} & \textbf{74.6} & \textbf{93.5} & \textbf{97.8} \\
    \hlineB{2}
    
    \multicolumn{1}{|c|}{\cellcolor[HTML]{99CCFF} Hypersim $\to$ NYUv2} & \cellcolor[HTML]{FAE5D3} MAE & \cellcolor[HTML]{FAE5D3} AbsRel & \cellcolor[HTML]{FAE5D3} RMSE & \cellcolor[HTML]{D5F5E3} $\delta_1$ & \cellcolor[HTML]{D5F5E3} $\delta_2$ & \cellcolor[HTML]{D5F5E3} $\delta_3$ \\
    \hline
      ConvNeXt-small & 0.434 & 0.165 & 0.598 & 75.7 & 94.3 & 98.5 \\
      ConvNeXt-small + Meta & \textbf{0.415} & \textbf{0.155} & \textbf{0.575} & \textbf{77.8} & \textbf{95.1} & \textbf{98.8} \\
      ConvNeXt-base & 0.396 & 0.150 & 0.549 & 79.6 & 95.6 & 98.9 \\
      ConvNeXt-base + Meta & \textbf{0.386} & \textbf{0.141} & \textbf{0.524} & \textbf{80.3} & \textbf{96.0} & \textbf{99.0} \\
      ConvNeXt-large & 0.389 & 0.149 & 0.542 & 79.8 & 95.6 & 98.8 \\
      ConvNeXt-large + Meta & \textbf{0.375} & \textbf{0.140} & \textbf{0.517} & \textbf{81.2} & \textbf{96.2} & \textbf{99.1} \\
    \hline
    \multicolumn{1}{|c|}{\cellcolor[HTML]{99CCFF} Hypersim $\to$ Replica} & \cellcolor[HTML]{FAE5D3} MAE & \cellcolor[HTML]{FAE5D3} AbsRel & \cellcolor[HTML]{FAE5D3} RMSE & \cellcolor[HTML]{D5F5E3} $\delta_1$ & \cellcolor[HTML]{D5F5E3} $\delta_2$ & \cellcolor[HTML]{D5F5E3} $\delta_3$ \\
    \hline
      ConvNeXt-small & 0.307 & 0.189 & 0.417 & 72.4 & 92.1 & \textbf{97.5} \\
      ConvNeXt-small + Meta & \textbf{0.294} & \textbf{0.178} & \textbf{0.404} & \textbf{74.5} & \textbf{92.7} & \textbf{97.5} \\
      ConvNeXt-base  & 0.312 & 0.185 & 0.429 & 74.1 & 92.6 & 97.4 \\ 
      ConvNeXt-base + Meta & \textbf{0.288} & \textbf{0.173} & \textbf{0.399} & \textbf{75.6} & \textbf{93.3} & \textbf{97.9} \\ 
      ConvNeXt-large & 0.285 & 0.172 & 0.394 & 75.8 & 93.2 & 97.7 \\
      ConvNeXt-large + Meta & \textbf{0.273} & \textbf{0.165} & \textbf{0.380} & \textbf{77.0} & \textbf{94.0} & \textbf{98.1} \\
    \hlineB{2}
    \hline
  \end{tabular}
  \vspace{-20pt}
\end{center}
\end{table}

We further experiment on recent high-performing architecture dedicated to depth estimation, including BTS \cite{lee2019big}, DPT (hybrid and large size) \cite{Ranftl2021}, and DepthFormer \cite{li2022depthformer}. Table \ref{table:cross-dataset-dedicated} displays the comparison and shows that meta-initialization consistently improves performance for zero-shot cross-dataset inference using the dedicated architectures, gearing higher generalizability for the existing models. 


\begin{table}[tb!]
\begin{center}
  \caption{\textbf{Cross-Dataset evaluation using dedicated depth estimation networks.} Our meta-initialization (+Meta) can be plugged into several methods to stably improve them.}
  \vspace{-4pt}
  \footnotesize
  \label{table:cross-dataset-dedicated}
  
  \begin{tabular}[c]
  {
  p{3.15cm}<{\arraybackslash}|
  p{0.5cm}<{\centering\arraybackslash}|
  p{0.65cm}<{\centering\arraybackslash}|
  p{0.55cm}<{\centering\arraybackslash}|
  p{0.35cm}<{\centering\arraybackslash}|
  p{0.35cm}<{\centering\arraybackslash}|
  p{0.35cm}<{\centering\arraybackslash}}
  \hlineB{2}
  
       \multicolumn{1}{|c|}{\cellcolor[HTML]{99CCFF} Hypersim $\to$ Replica} & \cellcolor[HTML]{FAE5D3} MAE & \cellcolor[HTML]{FAE5D3} AbsRel & \cellcolor[HTML]{FAE5D3} RMSE & \cellcolor[HTML]{D5F5E3} $\delta_1$ & \cellcolor[HTML]{D5F5E3} $\delta_2$ & \cellcolor[HTML]{D5F5E3} $\delta_3$ \\
    \hline
    BTS-ResNet50 \cite{lee2019big} & 0.365 & 0.226& 0.505 & 69.5 & 88.9 & 96.2 \\
    BTS-ResNet50 + Meta  & \textbf{0.328} & \textbf{0.208} & \textbf{0.475} & \textbf{70.6} & \textbf{90.0} & \textbf{96.5}\\
    BTS-ResNet101 \cite{lee2019big} & 0.339 & 0.214 & 0.488 & 69.9 & 89.8& 96.3\\
    BTS-ResNet101 + Meta & \textbf{0.322} & \textbf{0.200} & \textbf{0.469} & \textbf{71.0} & \textbf{90.4}& \textbf{96.8}\\
    \hline
    DepthFormer \cite{li2022depthformer} & 0.303 & 0.185 & 0.415 & 72.9 & 92.3 & 97.4 \\
      DepthFormer + Meta  & \textbf{0.292} & \textbf{0.180} & \textbf{0.400} & \textbf{74.3} & \textbf{92.7} & \textbf{97.5} \\
      \hline
      DPT-hybrid \cite{Ranftl2021} & 0.315 & 0.197 & 0.455 & 71.2 & 90.7 & 96.7 \\ 
      DPT-hybrid + Meta & \textbf{0.289} & \textbf{0.170} & \textbf{0.396} & \textbf{75.6} & \textbf{93.5} & \textbf{97.8} \\
      DPT-large \cite{Ranftl2021} & 0.294 & 0.172 & 0.401 & 75.4 & 93.3 & 97.6 \\
      DPT-large + Meta & \textbf{0.268} & \textbf{0.162} & \textbf{0.376} & \textbf{77.2} & \textbf{94.2} & \textbf{98.1} \\
    \hline
    AdaBins \cite{bhat2021adabins} & 0.331 & 0.210 & 0.445 & 70.2 & 90.1 & 96.6 \\
    AdaBins + Meta  & \textbf{0.319} & \textbf{0.198} & \textbf{0.432} & \textbf{71.7} & \textbf{92.0} & \textbf{97.3}\\
    \hline
    GLPDepth \cite{kim2022global} & 0.309 & 0.188 & 0.418 & 72.9 & 92.4& 97.6\\
    GLPDepth + Meta & \textbf{0.290} & \textbf{0.175} & \textbf{0.400} & \textbf{74.6} & \textbf{92.8}& \textbf{97.7}\\
    \hlineB{2}
    \hline
  
   \multicolumn{1}{|c|}{\cellcolor[HTML]{99CCFF} Hypersim $\to$ NYUv2} & \cellcolor[HTML]{FAE5D3} MAE & \cellcolor[HTML]{FAE5D3} AbsRel & \cellcolor[HTML]{FAE5D3} RMSE & \cellcolor[HTML]{D5F5E3} $\delta_1$ & \cellcolor[HTML]{D5F5E3} $\delta_2$ & \cellcolor[HTML]{D5F5E3} $\delta_3$ \\
    \hline
    BTS-ResNet50 \cite{lee2019big} & 0.487 & 0.196&0.654 & 71.8 &90.4 &95.6 \\
    BTS-ResNet50+Meta  & \textbf{0.455}& \textbf{0.178} & \textbf{0.628} & \textbf{73.9} & \textbf{92.4} & \textbf{97.3}\\
    BTS-ResNet101 \cite{lee2019big} & 0.468 & 0.187 & 0.641 & 72.3 & 90.8 & 95.8 \\
    BTS-ResNet101+Meta  & \textbf{0.450} & \textbf{0.175} & \textbf{0.623} & \textbf{74.2} & \textbf{92.6} & \textbf{97.5}\\
    \hline
      DepthFormer \cite{li2022depthformer} & 0.442 & 0.169 & 0.608 & 75.1 & 93.9 & \textbf{98.2} \\
      DepthFormer + Meta  & \textbf{0.416} & \textbf{0.157} & \textbf{0.580} & \textbf{77.8} & \textbf{94.3} & \textbf{98.2} \\
    \hline
      DPT-hybrid \cite{Ranftl2021} & 0.409 & 0.149 & 0.580 & 78.9 & 94.8 & 98.3 \\ 
      DPT-hybrid + Meta & \textbf{0.395} & \textbf{0.140} & \textbf{0.559} & \textbf{81.0} & \textbf{96.4} & \textbf{99.1} \\
      DPT-large \cite{Ranftl2021} & 0.373 & 0.136 & 0.534 & 82.3 & 96.2 & 98.8 \\
      DPT-large + Meta & \textbf{0.364} & \textbf{0.131} & \textbf{0.520} & \textbf{83.2} & \textbf{96.6} & \textbf{99.1} \\
    \hline
    AdaBins \cite{bhat2021adabins} & 0.469 & 0.188&0.642 & 72.6 &91.2 &96.6 \\
    AdaBins+Meta  & \textbf{0.448}& \textbf{0.175} & \textbf{0.625} & \textbf{74.0} & \textbf{92.6} & \textbf{97.4}\\
    \hline
      GLPDepth \cite{kim2022global} & 0.438 & 0.169 & 0.604 & 75.3 & 93.9 & 98.2 \\
      GLPDepth+Meta  & \textbf{0.414} & \textbf{0.158} & \textbf{0.583} & \textbf{77.9} & \textbf{94.3} & \textbf{98.3} \\
      
    \hlineB{2}
    \hline
  \end{tabular}
  \vspace{-19pt}
\end{center}
\end{table}

\subsection{Depth Supervision in NeRF}
\label{experiments:nerf}
\vspace{-2pt}
We show that more accurate depth from meta-initialization can better supervise the distance $d$ a ray travels in NeRF. $d$ is determined by the volumetric rendering rule \cite{deng2022depth}. Additional to pixel color loss, we use monocular predicted distance map $d^*$, converted from depth, to supervise the training by $\mathcal{L}_D=|d^*-d|$. The experiment is conducted on Replica's office-0 environment with 180 training views. After 30K training steps, we obtain NeRF-rendered views and calculate commonly-used image quality metrics (PSNR and SSIM, the higher the better). We use ConvNeXt-base to predict $d^*$. The comparison is made between using (A) without meta-initialization and (B) with meta-initialization.
Results: (A): (38.67, 0.9629), (B): (39.29, 0.9680). The results show better image quality is attained induced by better 3D representation in depth from meta-initialization. 
See more results in Supp Sec. F.

%% file: 05_conclusion.tex
\vspace{-3pt}
\section{Conclusion and Limitation}
\vspace{-3pt}
\label{conclusion}
We first validate single-stage meta-learning estimates coarse but smooth depth from global context (\ref{experiments:scene-fitting}). It is a better initialization for the following supervised learning to obtain higher model generalizability on intra-dataset and zero-shot cross-dataset evaluation (\ref{experiments:intra-dataset},  \ref{experiments:cross-dataset}), as well as a better 3D representation in NeRF training. (\ref{experiments:nerf}).

\noindent \textbf{From depth's perspective}, 
\begin{itemize}[leftmargin=*,topsep=-0pt,itemsep=-1.5ex]
\item this work provides a simple learning scheme to gain generalizability without needs of extra data, constraints, advanced loss, or module design; 
\item is easy to plug into general architecture or dedicated depth framework and show concrete improvements; 
\item proposes zero-shot cross-dataset protocol to attend to in-the-wild performance that most prior work overlook.
\end{itemize}

\noindent \textbf{From meta-learning's perspective}, 
\begin{itemize}[leftmargin=*,topsep=-0pt,itemsep=-1.5ex]
\item this work chooses the challenging single-image setting and meta-optimizes across environments, unlike prior works for online video adaptation that meta-optimizes within a sequence with access to multiple frames; 
\item proposes fine-grained task to overcome lacks of affinity in sparse and irrelevant sampled images. 
\item studies a complex single-image real-valued regression problem rather than widely-studied classification.
\end{itemize}

The work is at intersection of the two research fields and we hope it drives the dual-stream research development.

\textbf{Discussion: large foundation models}.
Those models require a large corpus of pretrained data and still need a finetune RGB-D dataset to adapt, and we only require an RGB-D set. If the size of finetune set is not large, a large model may easily overfit by overparameterization and show lower generalization. We experiment finetuning foundation models using ConvNeXt-XXLarge and ViT-L/14 and compare with our meta-initialization. They are CLIP weights first and further tuned on ImageNet22K. Finetuning foundation models does not win over our +Meta on Replica $\to$ VA and only show marginal gain on HM3D $\to$ VA.

\begin{table}[h]
\label{table:intra}
\begin{center}
  \vspace{-13pt}
  \ssmall
  \begin{tabular}[c]
  {
  p{1.4cm}<{\arraybackslash}
  p{2.5cm}<{\centering\arraybackslash}
  p{1.0cm}<{\centering\arraybackslash}
  p{2.3cm}<{\centering\arraybackslash}}
  \hlineB{2}
      AbsRel$\downarrow$ & ConvNeXt-XXLarge & ViT-L/14 & ConvNeXt-Base+Meta \\
    \hline
      Replica  $\to$VA & 0.437 & 0.434 & \textbf{0.430} \\
      HM3D  $\to$VA & \textbf{0.160}  & 0.162 & 0.163 \\
    \hlineB{1}
    \hline
  \end{tabular}

  \vspace{-25pt}
\end{center}
\end{table}

\textbf{Limitation} This work focuses on estimating depth from single images, which has wide applications in 3D-aware image synthesis or inpainting when an online depth estimator is needed. Other scopes of depth from images, such as stereo, multiview, video, or online learning is beyond our scope.  

%% file: egpaper_final.bbl
\begin{thebibliography}{100}\itemsep=-1pt

\bibitem{amac2022masksplit}
Mustafa~Sercan Amac, Ahmet Sencan, Bugra Baran, Nazli Ikizler-Cinbis, and
  Ramazan~Gokberk Cinbis.
\newblock Masksplit: Self-supervised meta-learning for few-shot semantic
  segmentation.
\newblock In {\em WACV}, 2022.

\bibitem{andrychowicz2016learning}
Marcin Andrychowicz, Misha Denil, Sergio Gomez, Matthew~W Hoffman, David Pfau,
  Tom Schaul, Brendan Shillingford, and Nando De~Freitas.
\newblock Learning to learn by gradient descent by gradient descent.
\newblock {\em NeurIPS}, 2016.

\bibitem{antoniou2019train}
Antreas Antoniou, Harri Edwards, and Amos Storkey.
\newblock How to train your maml.
\newblock In {\em ICLR}, 2019.

\bibitem{arpit2017closer}
Devansh Arpit, Stanis{\l}aw Jastrz{\k{e}}bski, Nicolas Ballas, David Krueger,
  Emmanuel Bengio, Maxinder~S Kanwal, Tegan Maharaj, Asja Fischer, Aaron
  Courville, Yoshua Bengio, et~al.
\newblock A closer look at memorization in deep networks.
\newblock In {\em ICML}, 2017.

\bibitem{bae2022irondepth}
Gwangbin Bae, Ignas Budvytis, and Roberto Cipolla.
\newblock Irondepth: Iterative refinement of single-view depth using surface
  normal and its uncertainty.
\newblock {\em arXiv preprint arXiv:2210.03676}, 2022.

\bibitem{bai2021person30k}
Yan Bai, Jile Jiao, Wang Ce, Jun Liu, Yihang Lou, Xuetao Feng, and Ling-Yu
  Duan.
\newblock Person30k: A dual-meta generalization network for person
  re-identification.
\newblock In {\em CVPR}, 2021.

\bibitem{balaji2018metareg}
Yogesh Balaji, Swami Sankaranarayanan, and Rama Chellappa.
\newblock Metareg: Towards domain generalization using meta-regularization.
\newblock {\em NeurIPS}, 2018.

\bibitem{bhat2021adabins}
Shariq~Farooq Bhat, Ibraheem Alhashim, and Peter Wonka.
\newblock Adabins: Depth estimation using adaptive bins.
\newblock In {\em CVPR}, 2021.

\bibitem{bhat2022localbins}
Shariq~Farooq Bhat, Ibraheem Alhashim, and Peter Wonka.
\newblock Localbins: Improving depth estimation by learning local
  distributions.
\newblock In {\em ECCV}, 2022.

\bibitem{zoedepth}
Shariq~Farooq Bhat, Reiner Birkl, Diana Wofk, Peter Wonka, and Matthias
  Müller.
\newblock Zoedepth: Zero-shot transfer by combining relative and metric depth,
  2023.

\bibitem{bian2021auto}
Jia-Wang Bian, Huangying Zhan, Naiyan Wang, Tat-Jun Chin, Chunhua Shen, and Ian
  Reid.
\newblock Auto-rectify network for unsupervised indoor depth estimation.
\newblock {\em TPAMI}, 2021.

\bibitem{bui2021exploiting}
Manh-Ha Bui, Toan Tran, Anh Tran, and Dinh Phung.
\newblock Exploiting domain-specific features to enhance domain generalization.
\newblock {\em NeurIPS}, 2021.

\bibitem{cao2019meta}
Zhiying Cao, Tengfei Zhang, Wenhui Diao, Yue Zhang, Xiaode Lyu, Kun Fu, and
  Xian Sun.
\newblock Meta-seg: A generalized meta-learning framework for multi-class
  few-shot semantic segmentation.
\newblock {\em IEEE Access}, 2019.

\bibitem{Chen2021S2RDepthNet}
Xiaotian Chen, Yuwang Wang, Xuejin Chen, and Wenjun Zeng.
\newblock S2r-depthnet: Learning a generalizable depth-specific structural
  representation.
\newblock In {\em CVPR}, 2021.

\bibitem{chen2021meta}
Yinbo Chen, Zhuang Liu, Huijuan Xu, Trevor Darrell, and Xiaolong Wang.
\newblock Meta-baseline: exploring simple meta-learning for few-shot learning.
\newblock In {\em ICCV}, 2021.

\bibitem{chua2021fine}
Kurtland Chua, Qi Lei, and Jason~D Lee.
\newblock How fine-tuning allows for effective meta-learning.
\newblock {\em NeurIPS}, 2021.

\bibitem{deng2022depth}
Kangle Deng, Andrew Liu, Jun-Yan Zhu, and Deva Ramanan.
\newblock Depth-supervised nerf: Fewer views and faster training for free.
\newblock In {\em CVPR}, 2022.

\bibitem{Dijk_2019_ICCV}
Tom~van Dijk and Guido~de Croon.
\newblock How do neural networks see depth in single images?
\newblock In {\em ICCV}, 2019.

\bibitem{dou2019domain}
Qi Dou, Daniel Coelho~de Castro, Konstantinos Kamnitsas, and Ben Glocker.
\newblock Domain generalization via model-agnostic learning of semantic
  features.
\newblock {\em NeurIPS}, 2019.

\bibitem{feldman2020neural}
Vitaly Feldman and Chiyuan Zhang.
\newblock What neural networks memorize and why: Discovering the long tail via
  influence estimation.
\newblock {\em NeurIPS}, 2020.

\bibitem{finn2017model}
Chelsea Finn, Pieter Abbeel, and Sergey Levine.
\newblock Model-agnostic meta-learning for fast adaptation of deep networks.
\newblock In {\em ICML}, 2017.

\bibitem{finn2019online}
Chelsea Finn, Aravind Rajeswaran, Sham Kakade, and Sergey Levine.
\newblock Online meta-learning.
\newblock In {\em ICML}, 2019.

\bibitem{flacco2012depth}
Fabrizio Flacco, Torsten Kr{\"o}ger, Alessandro De~Luca, and Oussama Khatib.
\newblock A depth space approach to human-robot collision avoidance.
\newblock In {\em ICRA}, 2012.

\bibitem{fu2018deep}
Huan Fu, Mingming Gong, Chaohui Wang, Kayhan Batmanghelich, and Dacheng Tao.
\newblock Deep ordinal regression network for monocular depth estimation.
\newblock In {\em CVPR}, 2018.

\bibitem{gao2022matters}
Ning Gao, Hanna Ziesche, Ngo~Anh Vien, Michael Volpp, and Gerhard Neumann.
\newblock What matters for meta-learning vision regression tasks?
\newblock In {\em CVPR}, 2022.

\bibitem{Godard_2019_ICCV}
Clement Godard, Oisin Mac~Aodha, Michael Firman, and Gabriel~J. Brostow.
\newblock Digging into self-supervised monocular depth estimation.
\newblock In {\em ICCV}, 2019.

\bibitem{gong2021cluster}
Rui Gong, Yuhua Chen, Danda~Pani Paudel, Yawei Li, Ajad Chhatkuli, Wen Li,
  Dengxin Dai, and Luc Van~Gool.
\newblock Cluster, split, fuse, and update: Meta-learning for open compound
  domain adaptive semantic segmentation.
\newblock In {\em CVPR}, 2021.

\bibitem{guizilini2021sparse}
Vitor Guizilini, Rares Ambrus, Wolfram Burgard, and Adrien Gaidon.
\newblock Sparse auxiliary networks for unified monocular depth prediction and
  completion.
\newblock In {\em CVPR}, 2021.

\bibitem{guo2021metacorrection}
Xiaoqing Guo, Chen Yang, Baopu Li, and Yixuan Yuan.
\newblock Metacorrection: Domain-aware meta loss correction for unsupervised
  domain adaptation in semantic segmentation.
\newblock In {\em CVPR}, 2021.

\bibitem{he2016deep}
Kaiming He, Xiangyu Zhang, Shaoqing Ren, and Jian Sun.
\newblock Deep residual learning for image recognition.
\newblock In {\em CVPR}, 2016.

\bibitem{he2019task}
Xu He, Jakub Sygnowski, Alexandre Galashov, Andrei~A Rusu, Yee~Whye Teh, and
  Razvan Pascanu.
\newblock Task agnostic continual learning via meta learning.
\newblock {\em arXiv preprint arXiv:1906.05201}, 2019.

\bibitem{hochreiter2001learning}
Sepp Hochreiter, A~Steven Younger, and Peter~R Conwell.
\newblock Learning to learn using gradient descent.
\newblock In {\em ICANN}, 2001.

\bibitem{Occlusion2018}
Aleksander Holynski and Johannes Kopf.
\newblock Fast depth densification for occlusion-aware augmented reality.
\newblock In {\em SIGGRAPH Asia}, 2018.

\bibitem{hospedales2021meta}
Timothy~M Hospedales, Antreas Antoniou, Paul Micaelli, and Amos~J Storkey.
\newblock Meta-learning in neural networks: A survey.
\newblock {\em IEEE Transactions on Pattern Analysis and Machine Intelligence},
  2021.

\bibitem{irshad2021hierarchical}
Muhammad~Zubair Irshad, Chih-Yao Ma, and Zsolt Kira.
\newblock Hierarchical cross-modal agent for robotics vision-and-language
  navigation.
\newblock In {\em ICRA}, 2021.

\bibitem{jamal2019task}
Muhammad~Abdullah Jamal and Guo-Jun Qi.
\newblock Task agnostic meta-learning for few-shot learning.
\newblock In {\em CVPR}, 2019.

\bibitem{ji2021monoindoor}
Pan Ji, Runze Li, Bir Bhanu, and Yi Xu.
\newblock Monoindoor: Towards good practice of self-supervised monocular depth
  estimation for indoor environments.
\newblock In {\em ICCV}, 2021.

\bibitem{jiang2021plnet}
Hualie Jiang, Laiyan Ding, Junjie Hu, and Rui Huang.
\newblock Plnet: Plane and line priors for unsupervised indoor depth
  estimation.
\newblock In {\em 3DV}, 2021.

\bibitem{jun2022depth}
Jinyoung Jun, Jae-Han Lee, Chul Lee, and Chang-Su Kim.
\newblock Depth map decomposition for monocular depth estimation.
\newblock 2022.

\bibitem{kim2022global}
Doyeon Kim, Woonghyun Ga, Pyungwhan Ahn, Donggyu Joo, Sehwan Chun, and Junmo
  Kim.
\newblock Global-local path networks for monocular depth estimation with
  vertical cutdepth.
\newblock {\em arXiv preprint arXiv:2201.07436}, 2022.

\bibitem{lee2019meta}
Hae~Beom Lee, Taewook Nam, Eunho Yang, and Sung~Ju Hwang.
\newblock Meta dropout: Learning to perturb latent features for generalization.
\newblock In {\em ICLR}, 2019.

\bibitem{lee2019big}
Jin~Han Lee, Myung-Kyu Han, Dong~Wook Ko, and Il~Hong Suh.
\newblock From big to small: Multi-scale local planar guidance for monocular
  depth estimation.
\newblock {\em arXiv preprint arXiv:1907.10326}, 2019.

\bibitem{lee2022pixel}
Yuan-Hao Lee, Fu-En Yang, and Yu-Chiang~Frank Wang.
\newblock A pixel-level meta-learner for weakly supervised few-shot semantic
  segmentation.
\newblock In {\em WACV}, 2022.

\bibitem{li2021structdepth}
Boying Li, Yuan Huang, Zeyu Liu, Danping Zou, and Wenxian Yu.
\newblock Structdepth: Leveraging the structural regularities for
  self-supervised indoor depth estimation.
\newblock In {\em ICCV}, 2021.

\bibitem{li2018learning}
Da Li, Yongxin Yang, Yi-Zhe Song, and Timothy~M Hospedales.
\newblock Learning to generalize: Meta-learning for domain generalization.
\newblock In {\em AAAI}, 2018.

\bibitem{li2022depthformer}
Zhenyu Li, Zehui Chen, Xianming Liu, and Junjun Jiang.
\newblock Depthformer: Exploiting long-range correlation and local information
  for accurate monocular depth estimation.
\newblock {\em arXiv preprint arXiv:2203.14211}, 2022.

\bibitem{li2022binsformer}
Zhenyu Li, Xuyang Wang, Xianming Liu, and Junjun Jiang.
\newblock Binsformer: Revisiting adaptive bins for monocular depth estimation.
\newblock 2022.

\bibitem{liuva}
Ce Liu, Suryansh Kumar, Shuhang Gu, Radu Timofte, and Luc Van~Gool.
\newblock Va-depthnet: A variational approach to single image depth prediction.
\newblock In {\em ICLR}, 2023.

\bibitem{liu2022convnet}
Zhuang Liu, Hanzi Mao, Chao-Yuan Wu, Christoph Feichtenhofer, Trevor Darrell,
  and Saining Xie.
\newblock A convnet for the 2020s.
\newblock {\em CVPR}, 2022.

\bibitem{loshchilov2018decoupled}
Ilya Loshchilov and Frank Hutter.
\newblock Decoupled weight decay regularization.
\newblock In {\em ICLR}, 2018.

\bibitem{luo2022towards}
Xinyu Luo, Jiaming Zhang, Kailun Yang, Alina Roitberg, Kunyu Peng, and Rainer
  Stiefelhagen.
\newblock Towards robust semantic segmentation of accident scenes via
  multi-source mixed sampling and meta-learning.
\newblock In {\em CVPRW}, 2022.

\bibitem{nascimento2020collision}
Hugo Nascimento, Martin Mujica, and Mourad Benoussaad.
\newblock Collision avoidance interaction between human and a hidden robot
  based on kinect and robot data fusion.
\newblock {\em IEEE Robotics and Automation Letters}, 6(1):88--94, 2020.

\bibitem{ni2021close}
Renkun Ni, Manli Shu, Hossein Souri, Micah Goldblum, and Tom Goldstein.
\newblock The close relationship between contrastive learning and
  meta-learning.
\newblock In {\em ICLR}, 2022.

\bibitem{nichol2018first}
Alex Nichol, Joshua Achiam, and John Schulman.
\newblock On first-order meta-learning algorithms.
\newblock {\em arXiv preprint arXiv:1803.02999}, 2018.

\bibitem{P3Depth}
Vaishakh Patil, Christos Sakaridis, Alexander Liniger, and Luc Van~Gool.
\newblock {P3Depth}: Monocular depth estimation with a piecewise planarity
  prior.
\newblock In {\em CVPR}, 2022.

\bibitem{qiao2020learning}
Fengchun Qiao, Long Zhao, and Xi Peng.
\newblock Learning to learn single domain generalization.
\newblock In {\em CVPR}, 2020.

\bibitem{rajendran2020meta}
Janarthanan Rajendran, Alexander Irpan, and Eric Jang.
\newblock Meta-learning requires meta-augmentation.
\newblock In {\em NeurIPS}, 2020.

\bibitem{rajeswaran2019meta}
Aravind Rajeswaran, Chelsea Finn, Sham~M Kakade, and Sergey Levine.
\newblock Meta-learning with implicit gradients.
\newblock {\em NeurIPS}, 2019.

\bibitem{ramakrishnan2021habitat}
Santhosh~K Ramakrishnan, Aaron Gokaslan, Erik Wijmans, Oleksandr Maksymets,
  Alex Clegg, John Turner, Eric Undersander, Wojciech Galuba, Andrew Westbury,
  Angel~X Chang, et~al.
\newblock Habitat-matterport {3D} dataset {(HM3D)}: 1000 large-scale {3D}
  environments for embodied {AI}.
\newblock {\em NeurIPS Datasets and Benchmarks Track}, 2021.

\bibitem{ramamonjisoa2019sharpnet}
Michael Ramamonjisoa and Vincent Lepetit.
\newblock Sharpnet: Fast and accurate recovery of occluding contours in
  monocular depth estimation.
\newblock {\em ICCVW}, 2019.

\bibitem{Ranftl2021}
Ren\'{e} Ranftl, Alexey Bochkovskiy, and Vladlen Koltun.
\newblock Vision transformers for dense prediction.
\newblock {\em ICCV}, 2021.

\bibitem{Ranftl2020}
Ren\'{e} Ranftl, Katrin Lasinger, David Hafner, Konrad Schindler, and Vladlen
  Koltun.
\newblock Towards robust monocular depth estimation: Mixing datasets for
  zero-shot cross-dataset transfer.
\newblock {\em TPAMI}, 2020.

\bibitem{roberts2020hypersim}
Mike Roberts and Nathan Paczan.
\newblock Hypersim: A photorealistic synthetic dataset for holistic indoor
  scene understanding.
\newblock {\em ICCV}, 2021.

\bibitem{roessle2022dense}
Barbara Roessle, Jonathan~T Barron, Ben Mildenhall, Pratul~P Srinivasan, and
  Matthias Nie{\ss}ner.
\newblock Dense depth priors for neural radiance fields from sparse input
  views.
\newblock In {\em CVPR}, 2022.

\bibitem{ruder2016overview}
Sebastian Ruder.
\newblock An overview of gradient descent optimization algorithms.
\newblock {\em arXiv preprint arXiv:1609.04747}, 2016.

\bibitem{schmidhuber1987evolutionary}
J{\"u}rgen Schmidhuber.
\newblock Evolutionary principles in self-referential learning, or on learning
  how to learn: the meta-meta-... hook.
\newblock {\em Technische Universit{\"a}t M{\"u}nchen, PhD thesis}, 1987.

\bibitem{schmidt2018grasping}
Philipp Schmidt, Nikolaus Vahrenkamp, Mirko W{\"a}chter, and Tamim Asfour.
\newblock Grasping of unknown objects using deep convolutional neural networks
  based on depth images.
\newblock In {\em ICRA}, 2018.

\bibitem{shu2021open}
Yang Shu, Zhangjie Cao, Chenyu Wang, Jianmin Wang, and Mingsheng Long.
\newblock Open domain generalization with domain-augmented meta-learning.
\newblock In {\em CVPR}, 2021.

\bibitem{silberman2012indoor}
Nathan Silberman, Derek Hoiem, Pushmeet Kohli, and Rob Fergus.
\newblock Indoor segmentation and support inference from rgbd images.
\newblock In {\em ECCV}, 2012.

\bibitem{straub2019replica}
Julian Straub, Thomas Whelan, Lingni Ma, Yufan Chen, Erik Wijmans, Simon Green,
  Jakob~J Engel, Raul Mur-Artal, Carl Ren, Shobhit Verma, et~al.
\newblock The replica dataset: A digital replica of indoor spaces.
\newblock {\em arXiv preprint arXiv:1906.05797}, 2019.

\bibitem{sun2022learn}
Qiyu Sun, Gary~G Yen, Yang Tang, and Chaoqiang Zhao.
\newblock Learn to adapt for monocular depth estimation.
\newblock {\em arXiv preprint arXiv:2203.14005}, 2022.

\bibitem{tai2018socially}
Lei Tai, Jingwei Zhang, Ming Liu, and Wolfram Burgard.
\newblock Socially compliant navigation through raw depth inputs with
  generative adversarial imitation learning.
\newblock In {\em ICRA}, 2018.

\bibitem{tan2022depth}
Sinan Tan, Mengmeng Ge, Di Guo, Huaping Liu, and Fuchun Sun.
\newblock Depth-aware vision-and-language navigation using scene query
  attention network.
\newblock In {\em ICRA}, 2022.

\bibitem{tian2020differentiable}
Pinzhuo Tian, Zhangkai Wu, Lei Qi, Lei Wang, Yinghuan Shi, and Yang Gao.
\newblock Differentiable meta-learning model for few-shot semantic
  segmentation.
\newblock In {\em AAAI}, 2020.

\bibitem{tonioni2019learning}
Alessio Tonioni, Oscar Rahnama, Thomas Joy, Luigi~Di Stefano, Thalaiyasingam
  Ajanthan, and Philip~HS Torr.
\newblock Learning to adapt for stereo.
\newblock In {\em CVPR}, 2019.

\bibitem{tonioni2019real}
Alessio Tonioni, Fabio Tosi, Matteo Poggi, Stefano Mattoccia, and Luigi~Di
  Stefano.
\newblock Real-time self-adaptive deep stereo.
\newblock In {\em CVPR}, 2019.

\bibitem{viereck2017learning}
Ulrich Viereck, Andreas Pas, Kate Saenko, and Robert Platt.
\newblock Learning a visuomotor controller for real world robotic grasping
  using simulated depth images.
\newblock In {\em CoRL}, 2017.

\bibitem{watson2019self}
Jamie Watson, Michael Firman, Gabriel~J Brostow, and Daniyar Turmukhambetov.
\newblock Self-supervised monocular depth hints.
\newblock In {\em ICCV}, 2019.

\bibitem{watson2021temporal}
Jamie Watson, Oisin Mac~Aodha, Victor Prisacariu, Gabriel Brostow, and Michael
  Firman.
\newblock The temporal opportunist: Self-supervised multi-frame monocular
  depth.
\newblock In {\em CVPR}, 2021.

\bibitem{wu2016occlusion}
Cho-Ying Wu and Jian-Jiun Ding.
\newblock Occlusion pattern-based dictionary for robust face recognition.
\newblock In {\em 2016 IEEE International Conference on Multimedia and Expo
  (ICME)}, 2016.

\bibitem{wu2018occluded}
Cho~Ying Wu and Jian~Jiun Ding.
\newblock Occluded face recognition using low-rank regression with generalized
  gradient direction.
\newblock {\em Pattern Recognition}, 80:256--268, 2018.

\bibitem{wu2019nonconvex}
Cho-Ying Wu and Jian-Jiun Ding.
\newblock Nonconvex approach for sparse and low-rank constrained models with
  dual momentum.
\newblock {\em arXiv preprint arXiv:1906.02433}, 2019.

\bibitem{wu2023inspacetype}
Cho-Ying Wu, Quankai Gao, Chin-Cheng Hsu, Te-Lin Wu, Jing-Wen Chen, and Ulrich
  Neumann.
\newblock Inspacetype: Reconsider space type in indoor monocular depth
  estimation.
\newblock {\em arXiv preprint arXiv:2309.13516}, 2023.

\bibitem{wu2022cross}
Cho-Ying Wu, Chin-Cheng Hsu, and Ulrich Neumann.
\newblock Cross-modal perceptionist: Can face geometry be gleaned from voices?
\newblock In {\em Proceedings of the IEEE/CVF Conference on Computer Vision and
  Pattern Recognition}, 2022.

\bibitem{wu2020geometry}
Cho-Ying Wu, Xiaoyan Hu, Michael Happold, Qiangeng Xu, and Ulrich Neumann.
\newblock Geometry-aware instance segmentation with disparity maps.
\newblock {\em arXiv preprint arXiv:2006.07802}, 2020.

\bibitem{wu2019efficient}
Cho~Ying Wu and Ulrich Neumann.
\newblock Efficient multi-domain dictionary learning with gans.
\newblock In {\em 2019 IEEE Global Conference on Signal and Information
  Processing (GlobalSIP)}, 2019.

\bibitem{wu2019salient}
Cho-Ying Wu and Ulrich Neumann.
\newblock Salient building outline enhancement and extraction using iterative
  l0 smoothing and line enhancing.
\newblock In {\em 2019 IEEE International Conference on Image Processing
  (ICIP)}, 2019.

\bibitem{wu2021scene}
Cho-Ying Wu and Ulrich Neumann.
\newblock Scene completeness-aware lidar depth completion for driving scenario.
\newblock In {\em ICASSP}, 2021.

\bibitem{wu2022toward}
Cho-Ying Wu, Jialiang Wang, Michael Hall, Ulrich Neumann, and Shuochen Su.
\newblock Toward practical monocular indoor depth estimation.
\newblock In {\em CVPR}, 2022.

\bibitem{wu2021synergy}
Cho-Ying Wu, Qiangeng Xu, and Ulrich Neumann.
\newblock Synergy between 3dmm and 3d landmarks for accurate 3d facial
  geometry.
\newblock In {\em 2021 International Conference on 3D Vision (3DV)}, 2021.

\bibitem{yao2021improving}
Huaxiu Yao, Long-Kai Huang, Linjun Zhang, Ying Wei, Li Tian, James Zou, Junzhou
  Huang, et~al.
\newblock Improving generalization in meta-learning via task augmentation.
\newblock In {\em ICML}, 2021.

\bibitem{yin2020meta}
Mingzhang Yin, George Tucker, Mingyuan Zhou, Sergey Levine, and Chelsea Finn.
\newblock Meta-learning without memorization.
\newblock In {\em ICLR}, 2020.

\bibitem{yin2019enforcing}
Wei Yin, Yifan Liu, Chunhua Shen, and Youliang Yan.
\newblock Enforcing geometric constraints of virtual normal for depth
  prediction.
\newblock In {\em ICCV}, 2019.

\bibitem{yin2021learning}
Wei Yin, Jianming Zhang, Oliver Wang, Simon Niklaus, Long Mai, Simon Chen, and
  Chunhua Shen.
\newblock Learning to recover 3d scene shape from a single image.
\newblock In {\em CVPR}, 2021.

\bibitem{yu2020p}
Zehao Yu, Lei Jin, and Shenghua Gao.
\newblock P$^2$net: Patch-match and plane-regularization for unsupervised
  indoor depth estimation.
\newblock In {\em ECCV}, 2020.

\bibitem{yuan2022new}
Weihao Yuan, Xiaodong Gu, Zuozhuo Dai, Siyu Zhu, and Ping Tan.
\newblock New crfs: Neural window fully-connected crfs for monocular depth
  estimation.
\newblock {\em CVPR}, 2022.

\bibitem{zhang2019online}
Zhenyu Zhang, St{\'e}phane Lathuiliere, Andrea Pilzer, Nicu Sebe, Elisa Ricci,
  and Jian Yang.
\newblock Online adaptation through meta-learning for stereo depth estimation.
\newblock {\em arXiv preprint arXiv:1904.08462}, 2019.

\bibitem{zhang2020online}
Zhenyu Zhang, Stephane Lathuiliere, Elisa Ricci, Nicu Sebe, Yan Yan, and Jian
  Yang.
\newblock Online depth learning against forgetting in monocular videos.
\newblock In {\em CVPR}, 2020.

\bibitem{zhao2021learning}
Yuyang Zhao, Zhun Zhong, Fengxiang Yang, Zhiming Luo, Yaojin Lin, Shaozi Li,
  and Nicu Sebe.
\newblock Learning to generalize unseen domains via memory-based multi-source
  meta-learning for person re-identification.
\newblock In {\em CVPR}, 2021.

\bibitem{NEURIPS2019_e2c61965}
Yiqi Zhong, Cho-Ying Wu, Suya You, and Ulrich Neumann.
\newblock Deep rgb-d canonical correlation analysis for sparse depth
  completion.
\newblock In {\em Advances in Neural Information Processing Systems},
  volume~32, 2019.

\bibitem{zhou2019moving}
Junsheng Zhou, Yuwang Wang, Kaihuai Qin, and Wenjun Zeng.
\newblock Moving indoor: Unsupervised video depth learning in challenging
  environments.
\newblock In {\em CVPR}, 2019.

\end{thebibliography}
